\documentclass{article}

\usepackage{microtype}
\usepackage{graphicx}
\usepackage{subfigure}
\usepackage{booktabs} 

\usepackage{hyperref}

\usepackage[accepted]{icml2024}

\usepackage{amsmath}
\usepackage{amssymb}
\usepackage{mathtools}
\usepackage{amsthm}
\usepackage{multirow}
\usepackage{soul}
\usepackage{longtable}

\usepackage{xcolor}
\definecolor{darkgreen}{rgb}{0.0, 0.5, 0.0} 

\usepackage{algorithmic}
\usepackage[capitalize,noabbrev]{cleveref}
\usepackage{titlesec}
\titlespacing\section{0pt}{8pt}{0pt}
\titlespacing\subsection{0pt}{3pt}{0pt}
\titlespacing\paragraph{0pt}{0pt}{0pt}
\titlespacing\bfsection{0pt}{2pt}{0pt}

\theoremstyle{plain}
\newtheorem{theorem}{Theorem}[section]

\theoremstyle{definition}

\theoremstyle{remark}

\usepackage[textsize=tiny]{todonotes}
\usepackage[font=small,skip=6pt]{caption}

\newcommand{\bfsection}[1]{\noindent\textbf{#1}}

\newcommand{\lvlm}{$\mathcal{M}_\theta^{\text{LVLM}}$}

\newcommand{\comt}[1]{#1}

\renewcommand{\comt}[1]{}

\newcommand{\tabref}[1]{Table~\ref{#1}}
\newcommand{\figref}[1]{Fig.~\ref{#1}}
\newcommand{\eqnref}[1]{\text{Eq.}~(\ref{#1})}
\newcommand{\secref}[1]{\S\ref{#1}}
\newcommand{\appref}[1]{Appendix~\ref{#1}}

\icmltitlerunning{HALC: Object Hallucination Reduction via Adaptive Focal-Contrast Decoding}

\begin{document}

\twocolumn[
\icmltitle{HALC: Object Hallucination Reduction via Adaptive Focal-Contrast Decoding}



\icmlsetsymbol{equal}{*}

\begin{icmlauthorlist}
\icmlauthor{Zhaorun Chen}{equal,uchi}
\icmlauthor{Zhuokai Zhao}{equal,uchi}
\icmlauthor{Hongyin Luo}{mit}
\icmlauthor{Huaxiu Yao}{unc}
\icmlauthor{Bo Li}{uchi,uiuc}
\icmlauthor{Jiawei Zhou}{ttic}
\end{icmlauthorlist}

\icmlaffiliation{uchi}{University of Chicago, Chicago IL, USA}
\icmlaffiliation{uiuc}{University of Illinois at Urbana-Champaign, Champaign IL, USA}
\icmlaffiliation{mit}{Massachusetts Institute of Technology, Boston MA, USA}
\icmlaffiliation{unc}{UNC-Chapel Hill, Chapel Hill NC, USA}
\icmlaffiliation{ttic}{Toyota Technological Institute at Chicago, Chicago IL, USA}

\icmlcorrespondingauthor{Zhaorun Chen, Zhuokai Zhao, Bo Li}{\{zhaorun, zhuokai, bol\}@uchicago.edu}
\icmlcorrespondingauthor{Jiawei Zhou}{jzhou@ttic.edu}

\icmlkeywords{Machine Learning, ICML}

\vskip 0.3in
]



\printAffiliationsAndNotice{\icmlEqualContribution} 

\begin{abstract}
While large vision-language models (LVLMs) have demonstrated impressive capabilities in 
interpreting multi-modal contexts, they invariably suffer from object hallucinations (OH). 
%
We introduce \textbf{HALC}, a novel decoding algorithm designed to 
mitigate OH in LVLMs. 
HALC leverages distinct fine-grained optimal visual information in vision-language tasks and 
operates on both local and global contexts simultaneously.
Specifically, HALC integrates a robust auto-focal grounding mechanism (locally) to correct hallucinated tokens on the fly, and 
a specialized beam search algorithm (globally) to significantly reduce 
OH while preserving text generation quality.
%
Additionally, HALC can be integrated into any LVLMs as a plug-and-play module without
extra training.
Extensive experimental studies demonstrate the effectiveness of HALC in reducing OH, 
outperforming state-of-the-arts across four benchmarks.
%
Code is released at \url{https://github.com/BillChan226/HALC}.
\end{abstract}

\section{Introduction}\label{sec:introduction}
The confluence of natural language processing (NLP) and computer vision (CV) has 
undergone a transformative shift over the past years with the introduction of 
vision-language models (VLMs)~\cite{zhu2023minigpt, liu2023visual, zhang2024rankclip}.
Although VLMs have shown exceptional proficiency in integrating and interpreting intricate 
data across both textual and visual modalities,
%
a significant challenge emerged as the 
phenomenon of \textit{object hallucination (OH)}, where VLMs erroneously generate 
hallucinated objects and descriptions
within their outputs~\cite{rohrbach2018object}.
Based on the different parts of the sentences that are being hallucinated,
OH can be categorized into three types: 
object \textit{existence}, \textit{attribute}, and \textit{relationship} hallucinations~\cite{gunjal2023detecting, zhai2023halle}.
%

OH has been a persistent challenge since the earlier stages of the VLM 
development~\cite{rohrbach2018object}.
And it has been gaining increased attention, especially when recent research indicates 
that even the much more sophisticated and capable large vision-language models (LVLMs) 
are not immune to it~\cite{dai2022plausible, li2023evaluating, guan2023hallusionbench}.
Numerous efforts have been devoted to mitigating OH in the context of
LVLMs, including a post-hoc approach that corrects the LVLM output after 
completion~\cite{zhou2023analyzing}, 
a self-correction pipeline for OH mitigation~\cite{yin2023woodpecker}, 
and various decoding strategies that are tailored towards reducing OH via better 
textual or visual priors utilization~\cite{huang2023opera, leng2023mitigating}.

Despite the efforts, these approaches are not yet fully satisfying in terms of
eliminating OH.
More importantly, they mainly focus on mitigating object existence hallucination,
while assuming the attribute- and relationship-level hallucinations can be consequently
corrected through autoregressive decoding.
Furthermore, their reliance on more powerful external LVLMs~\cite{yin2023woodpecker}, 
repeated processing~\cite{zhou2023analyzing} or additional 
data~\cite{gunjal2023detecting} complicates their adaptations to existing LVLMs and
restricts their use cases.
The importance of OH reduction combined with the limitations in existing methods 
underscore the urgent need for developing novel approaches.

To this end, we introduce Object \textbf{H}allucination Reduction 
through \textbf{A}daptive Foca\textbf{L}-\textbf{C}ontrast decoding (\textbf{HALC}), 
a novel decoding strategy designed to effectively counter OH and can be easily integrated 
into any open-source LVLMs such as MiniGPT-4~\cite{chen2023minigpt}, 
LLaVA~\cite{liu2023visual} and mPLUG-Owl2~\cite{ye2023mplug}.
HALC addresses all three types of OH (existence, attribute, and relationship) 
while preserving generation quality in both local and global levels;
locally, it employs an \textit{adaptive focal-contrast grounding} mechanism to locate
the fine-grained optimal visual information to correct each generated token that might be 
hallucinating; 
and globally, it incorporates a \textit{matching-based beam search} that utilizes 
a visual matching score to steer the generation of the final outputs to balance both 
OH mitigation and text generation quality.

The main contributions of this paper are: (1) HALC, a novel, plug-and-play decoding 
algorithm that significantly reduces OH in LVLMs while preserving outputs generation 
quality;
(2) an open-sourced platform that unifies all major OH reduction 
baselines and state-of-the-arts (SOTAs)~\cite{chuang2023dola, zhou2023analyzing, yin2023woodpecker, huang2023opera, leng2023mitigating}, including HALC, into one framework 
providing convenient evaluations supporting major LVLM 
backbones~\cite{zhu2023minigpt, chen2023minigpt, liu2023visual, Dai2023InstructBLIPTG} 
and OH benchmarks and evaluation
metrics~\cite{rohrbach2018object, fu2023mme, li2023evaluating, liu2023improved}; and (3) comprehensive experimental studies that thoroughly evaluates HALC, demonstrating 
its superior capability in OH reduction over existing approaches.
%


\section{Related Work}\label{sec:related_work}
\bfsection{OH and its assessment.}
OH refers to the phenomenon where vision-language models (VLMs), including
both earlier BERT-based models~\cite{li2019visualbert, radford2021learning} and more 
recent LVLMs~\cite{liu2023visual, zhu2023minigpt,tu2023many,cui2023holistic,wang2024mementos, zhou2024calibrated}, erroneously generate unfaithful contents.
More specifically,~\citet{gunjal2023detecting} and~\citet{zhai2023halle} 
proposed that OH could be categorized into three types:
object \textit{existence} hallucination for 
the creation of non-existent objects, object \textit{attribute} hallucination for 
providing misleading descriptions, and object \textit{relationship} hallucination for 
depicting incorrect inter-object 
relationships.
%

The most well-adopted metric specifically designed to evaluate OH is 
CHAIR~\cite{rohrbach2018object}, which was motivated after~\citet{rohrbach2018object} 
discovered that existing metrics that measure the output's text quality, 
such as CIDEr~\cite{vedantam2015cider}, is misleading at representing hallucinations
(higher CIDEr score may correlate with higher OH).
Another notable and more recent metric is POPE~\cite{li2023evaluating}, which transforms the 
assessment of OH into a binary classification problem where metrics such as precision, recall
and accuracy are used to represent the level of OH.
In our evaluations, we utilize CHAIR and propose a new metric based on POPE, 
named \textit{OPOPE}, for thorough assessments of OH, while keeping
the standard text generation quality metrics such as BLEU~\cite{papineni2002bleu}, as an 
additional indicator to make sure little sacrifice in quality was made when mitigating OH.
%

\bfsection{Challenges and existing approaches.}
OH has been a persistent challenge over the past years~\cite{rohrbach2018object}. 
Despite numerous advancements in 
LVLMs~\cite{dai2022plausible, li2023evaluating,zhou2024aligning}, none of them can
produce faithful outputs without suffering from some level of OH.
Various strategies have been developed to this matter. 
For instance,~\citet{zhou2023analyzing} and~\citet{yin2023woodpecker} proposed post-hoc 
and self-correction pipelines, respectively. 
\citet{huang2023opera} and~\citet{leng2023mitigating} developed decoding strategies 
emphasizing better prior utilization. 
While effective, these approaches often require powerful external LVLMs or additional data, 
limiting their adaptability.
%

%
Distinct from these methods, HALC offers a novel decoding strategy that effectively 
reduces OH without necessitating extra LVLMs, training, or data. 
Integrating a novel adaptive focal-contrast grounding mechanism, HALC addresses both local and 
global contexts in OH reduction. 
Its compatibility with open-source LVLMs like MiniGPT-4~\cite{zhu2023minigpt} and 
LLaVA~\cite{liu2023visual} further enhances its applicability.
And as previous approaches often study the problem under different settings and metrics~\cite{zhou2023analyzing, yin2023woodpecker, huang2023opera, leng2023mitigating}, to promote the development of OH reduction in general, we implement an open-source 
platform which hosts both the proposed HALC and other methods, supporting various LVLM 
backbones and evaluation metrics.
%


\section{Background and Motivation}
\label{sec:background}


\subsection{Problem Formulation}

We consider an LVLM {\lvlm} parameterized by $\theta$, with a 
general architecture consisting of a vision encoder, a vision-text interface module, 
and a text decoder. 
For an image-grounded text generation task, given a textual query $x$ and an input image $v$, 
$v$ is first processed by the vision encoder into a visual embedding, then transformed by the 
interface module as the input to the text decoder together with the query $x$, and finally 
decoded into a textual response $y$ autoregressively. 
Formally, we have
\begin{equation}
y_t \sim p_{\theta}(\cdot | v, x, y_{<t}) \propto \exp f_{\theta}(\cdot | v, x, y_{<t})
\end{equation}
where $y_t$ denotes the $t^{th}$ token, $y_{<t}$ is the token sequence generated up to 
time step $t$, and $f_{\theta}$ is the logit distribution (unnormalized log-probabilities) 
produced by {\lvlm}.

OH happens when some parts of the text generation $y$ conflicts with the 
input image $v$. 
The goal of OH reduction is to minimize the occurrence of hallucination 
tokens and preserve the faithfulness to $v$ when addressing the query $x$, while maintaining 
a high-quality generation of text $y$.
%



\subsection{Why Does OH Occur?}
OH in VLMs can be attributed to various factors, including but not limited to the 
inherent biases in the training data caused by 
co-occurrence~\cite{biten2022let, zhou2023analyzing}, visual 
uncertainty due to model's statistical bias and priors~\cite{leng2023mitigating}, as
well as the limitations in current models' ability to discern context and fact
accurately during the entire output generation process~\cite{daunhawer2021limitations}.
Studies have also shown that OH is not random but exhibits certain
patterns and dependencies, such as its 
co-existence with knowledge aggregation pattern~\cite{huang2023opera}, and the tendency to 
occur with objects positioned later in the generated descriptions~\cite{zhou2023analyzing}.

A closer examination of these analysis suggests that the autoregressive nature of the
LVLMs may be a fundamental factor contributing to their hallucinatory behaviors.
Specifically, autoregressive decoding makes LVLMs progressively rely more on 
textual information including both the query $x$ and the increasing history generations 
$y_{<t}$, while unavoidably reducing reliance on the visual input.
This imbalance results in a significant deviation from accurate representation of the
visual input, ultimately culminating in OH with behaviors and patterns observed in the 
aforementioned studies~\cite{zhou2023analyzing, leng2023mitigating}.
This is especially obvious when longer responses are generated, which explains the 
correlation between higher OH and larger maximum token lengths, as seen 
in~\citet{huang2023opera}.
%

\subsection{Fine-grained Visual Knowledge Reduces OH}
\begin{figure*}[t]
    \centering
    \includegraphics[width=0.85\textwidth]{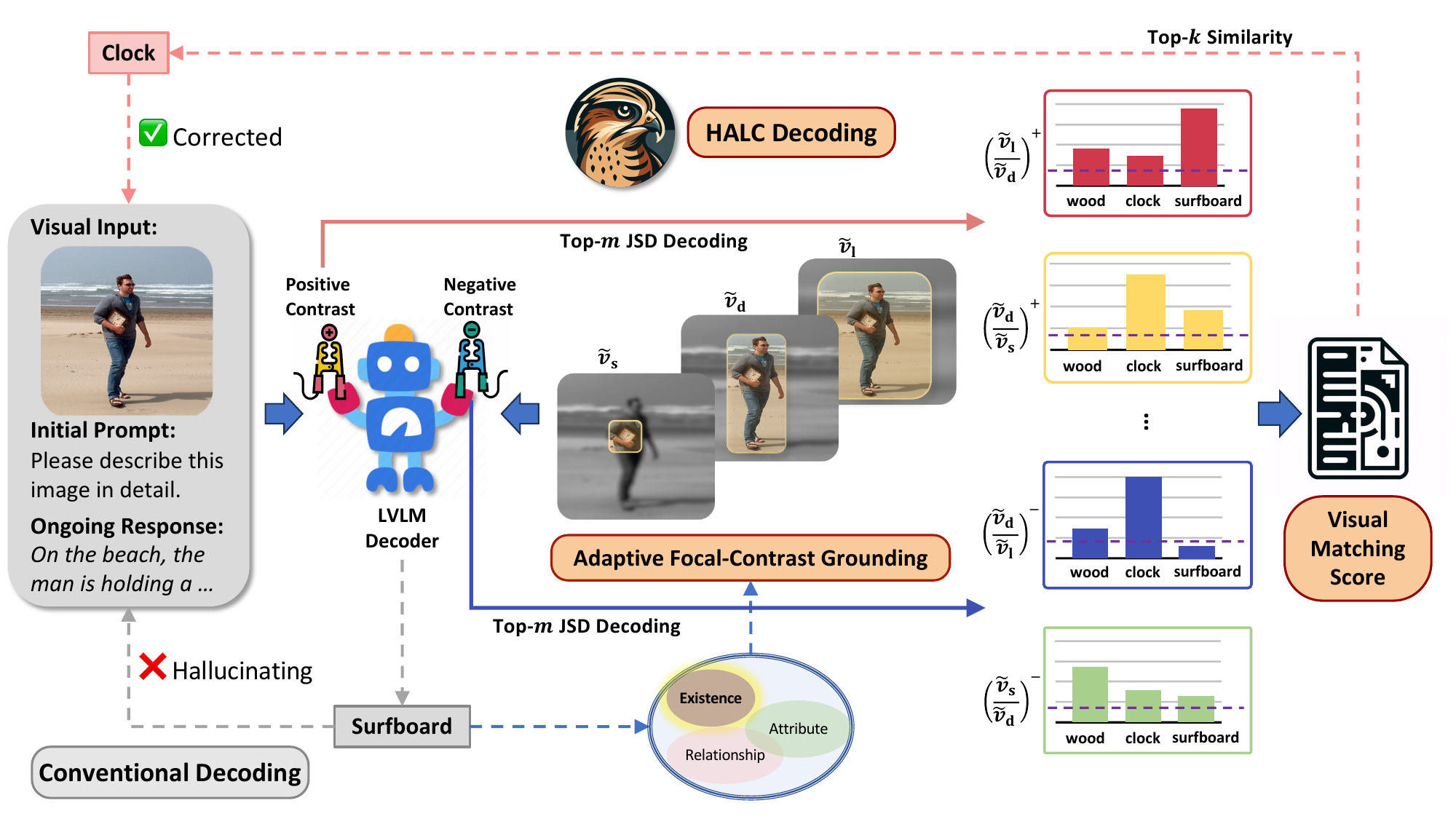}
    \caption{
    An overview of HALC. 
    As LVLM autoregressively generates texts  w.r.t. an image input (e.g. a man holding a clock on the beach), the conventional decoding 
    method may hallucinate the \textit{clock} as \textit{surfboard}. However, HALC 
    corrects this potential hallucination by first locating its 
    visual grounding $v_d$, then sample $n$ distinctive yet overlapping FOVs (e.g. $\tilde{v}_s$, $\tilde{v}_d$, $\tilde{v}_l$).
    Next, all FOVs are fed back into the 
    LVLM, along with the current ongoing response, obtaining $n$ logits distributions.
    Then we compute Jensen-Shannon Divergence (JSD) between each pair of the 
    $n$ distributions, and select the top $m$ pairs,
    %
    providing $2m$ next-token candidates by bi-directional contrasted logits distributions.
    Each of the $2m$ candidates are then appended to the $k$ ongoing beams (beam search omitted in the figure for simplicity), resulting 
    in $2mk$ response candidates.
    Finally, $k$ best responses are selected according to the global visual matching score between current text and original image,
    completing the current decoding 
    round with the hallucinating token \textit{surfboard} successfully corrected to 
    \textit{clock}.
    }
    \label{fig:halc_overview}
    \vspace{-0.15in}
\end{figure*}

To mitigate the disproportionate reliance on the textual and visual information during the 
autoregressive text generation, the process can be enhanced by continuously incorporating 
targeted visual information.
As faithful text generations should guarantee that object-related text tokens are well grounded 
in the visual input, we hypothesize that the generation can benefit from focusing more on the
\textit{fine-grained visual context} for different object-related tokens.
For example, for an image showing \textit{a man holding a clock on the beach} as in 
Fig.~\ref{fig:halc_overview}, the generation of the \textit{clock} token can be well grounded in a 
smaller region of the image, which we call a specific \textit{visual context}, ideally excluding the 
beach which is distracting.
Therefore, our key insight in mitigating OH lies in identifying a token-wise optimal visual 
context to provide the most informative visual grounding while decoding a specific token.

\begin{figure}[t]
    \centering
    \includegraphics[width=0.9\linewidth]{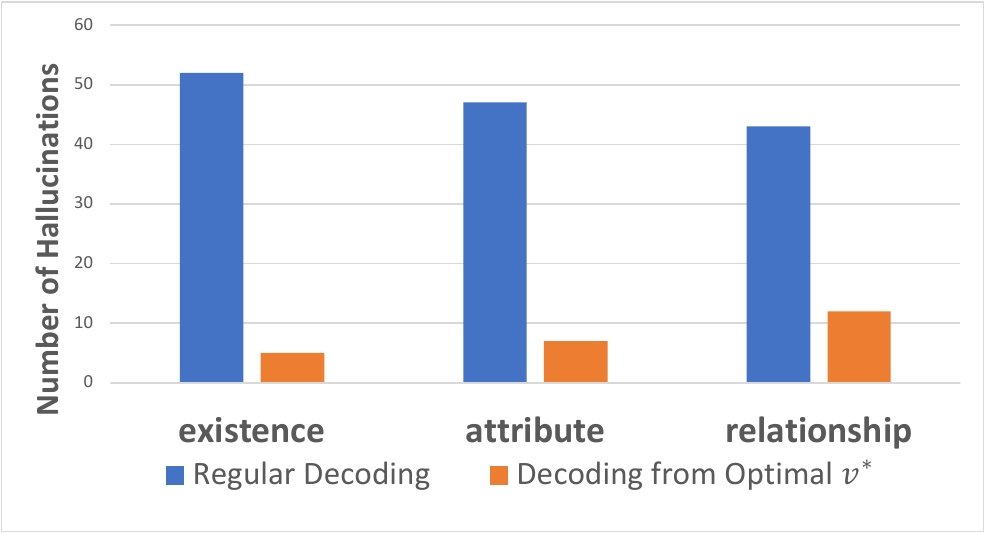}
    \caption{
    On average, over 84.5\% of the observed existence, attribute, and relationship 
    hallucinations are reduced by leveraging some optimal visual context $v^*$.
    Blue bar denotes number of hallucinated tokens on each corresponding MME sub-task,  
    while orange bar denotes results when decoding from the oracle $v^*$.
    }
    \label{fig:pattern_analysis}
    \vspace{-0.25in}
\end{figure}

We verify our hypothesis through an empirical pilot study. 
\figref{fig:pattern_analysis} shows the oracle performance of OH levels when we 
rely on optimal visual contexts for tokens through brute-force search, with greedy decoding 
on the MME benchmark~\cite{fu2023mme} on three categories of 
OH.\footnote{Details of this oracle analysis can be found in Appendix~\ref{app:contexts}.}
We can see that for most cases, there are optimal visual contexts where decoding from them 
eliminates over 84.5\% of the hallucinations.
This motivates our approach of identifying different visual contexts for object-related token 
generations through \textit{adaptive focal-contrast decoding}, which is introduced in 
detail in the next section.

\section{Methodology}\label{sec:method}

An overview of the proposed HALC method is shown in \figref{fig:halc_overview}.
It operates at the token level during generation, with reliance on fine-grained visual information represented by samples of different visual contexts. By recomputing the token distributions from different visual context inputs and contrasting them, object-related token probabilities are redistributed to reduce hallucinations dynamically within the generation steps.
We describe the full procedures below.


\subsection{Object-related Token Identification}
\label{subsec:token_identify}

To focus on the most-probable hallucination sources and optimize time efficiency, we first identify tokens that are related to objects to be processed by HALC.
In particular, at each generation step $t$, we acquire the part-of-speech (POS) tag~\cite{honnibal2017spacy}\footnote{We use the small-sized spaCy English pipeline (\url{https://spacy.io/models/en}) for tagging each complete word.} 
of the currently generated token from the model {\lvlm}. If the token belongs to noun, 
adjective/adverb/number/verb/pronoun, or preposition, which correspond to object existence, 
attribute, and relationship hallucinations, respectively, we redo the current token generation with HALC.
For example, as seen in \figref{fig:halc_overview}, the newly generated token 
\textit{surfboard} is identified as it may contribute to the object existence hallucination.
Notice that we do not make any assumptions on whether or not the current token is hallucinating,
instead, we only determine if the token \textit{can} be prune to hallucination solely based on its 
syntactic category.
%


\subsection{Visual Context Retrieval}
\label{subsec:visual_retrieval}

To identify the fine-grained visual information for the current token, we first retrieve a visual context window $v_d=(w_d, h_d, p_d)$ corresponding to the token, where $w_d$ and $h_d$ are the width and height of the visual window, and $p_d$ is the center point.
Specifically, we employ a zero-shot detector $\mathcal{G}_{d}$ such as Grounding 
DINO~\cite{liu2023grounding} or OWLv2~\cite{minderer2023scaling} to locate the token 
within the original image input $v$.
Notably, despite the most common use case of these zero-shot detectors is to locate objects, 
they are trained to also provide good visual reference for adjective or prepositional phrase.
This is because during pre-training, the objective of these detection models is to associate 
words in text descriptions with specific regions in images~\cite{liu2023grounding}, 
which naturally includes attributes and relationships besides names.

Interestingly, we find that although the current token may technically be
non-existing when it represents a hallucination (e.g., \textit{surfboard} in 
\figref{fig:halc_overview}), it can still be accurately located by the detector in practice,
especially when the detector confidence threshold is set to lower values. 
%
%


\subsection{Adaptive Focal-contrast Grounding}
\label{subsec:logits_redistribution}
\begin{figure}[t]
    \centering
    \includegraphics[width=0.9\linewidth]{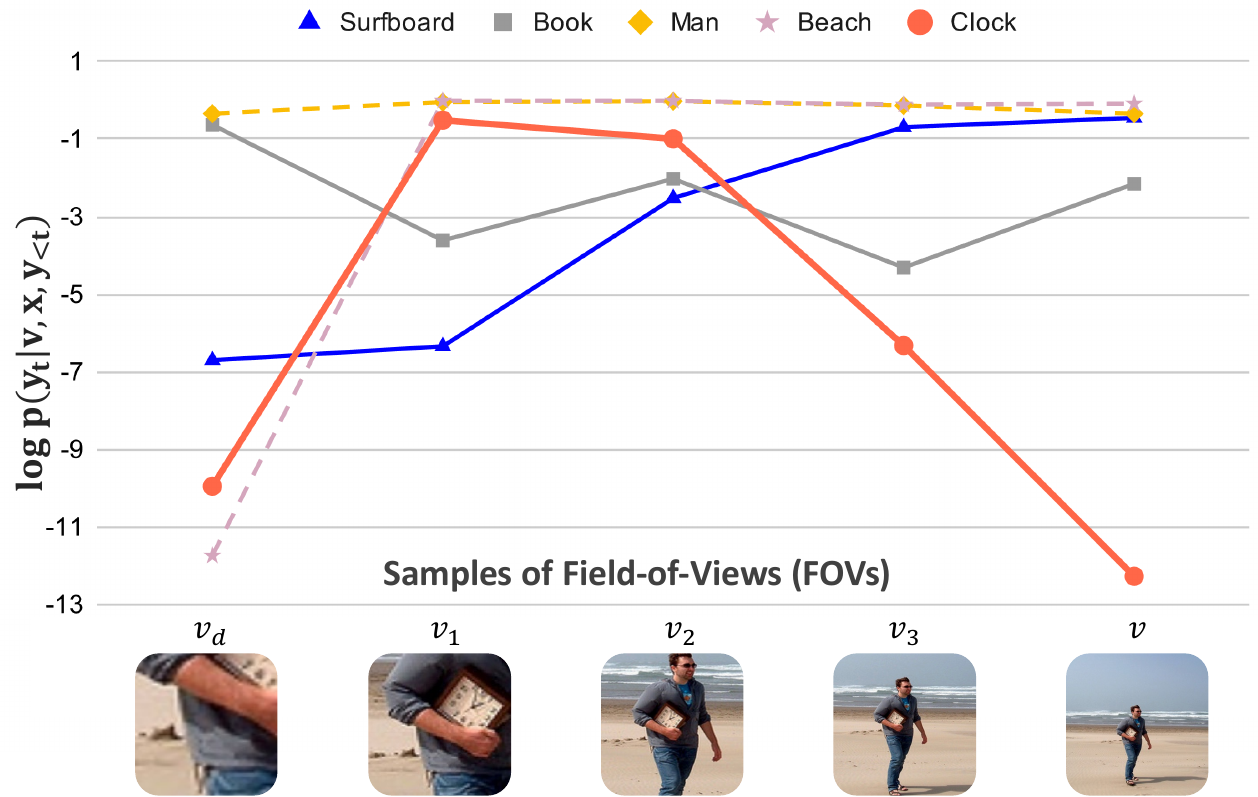}
    \caption{
    Log-likelihood of object tokens w.r.t. visual context samples in the FOV space, at the generation step in the example of \figref{fig:halc_overview}.
    Exponentially expanding FOVs are adopted. 
    While obvious objects (e.g. \textit{beach}, \textit{man}) are stable with high 
    likelihood, hallucinating objects are either noisy (e.g. \textit{book}) or shift 
    gradually with the context (e.g. \textit{surfboard}). 
    The victim token (e.g. \textit{clock}) usually display a drastically peaking pattern (local maximum).
    }\label{fig:optimal_visual_context}
    \vspace{-0.25in}
\end{figure}

While off-the-shelf detectors establish a meaningful reference $v_d$ within the original image 
input $v$, it is often not the optimal visual context for decoding.
In \figref{fig:optimal_visual_context}, we show an example of how token probabilities representing different objects change with different visual context windows, or field of views (FOVs) input to the vision model in {\lvlm}.
In this generation step, the ground-truth token ``clock'' (we call a \textit{victim token}) is hallucinated to ``surfboard''.
Although direct decoding from $v_d$ does not correct the hallucination as the probability of ``clock'' is still low, we can see that there exists a better visual context window $v_1$ that can correct the hallucination, and the curve corresponding to the faithful token ``clock'' displays a drastically peaking pattern.
This is a sharp difference from the patterns of other tokens, which display smaller contrasts when the visual contexts vary.
This observation motivates our approach of \textit{focal-contrast grounding} to adaptively adjust the object-related token probabilities, by sampling and selecting a range of most contrasting FOVs based on their decoding probabilities to best approximate the optimal visual contexts.


\bfsection{FOV sampling}.
We first sample a sequence of $n$ FOVs, $v_1, v_2, \ldots, v_n$, based on the initial visual 
context $v_d$.
There could be different approaches to come up with different FOVs conditioning on $v_d$.
To attain a larger coverage of the input image quickly, one strategy to sample FOVs is 
through an exponential expanding function, by setting 
\begin{equation}\label{eq:fov_expand}
    v_i=(w_i, h_i, p_i)=\left((1+\lambda)^i w_d, (1+\lambda)^i h_d, p_d\right)
\end{equation}
where $w_i, h_i, p_i$ are the width, height, and center of FOV $v_i$.
%

\bfsection{Dynamic visual context selection.}
Based on the observation from \figref{fig:optimal_visual_context}, we now select a set of 
FOVs based on a contrastive criterion in the text decoding space to better approximate the 
optimal visual context for the current token.
In particular, after obtaining $n$ different FOVs, we feed these visual contexts back into 
the model\footnote{We directly feed the cropped image to the FOV in the model.} {\lvlm}, resulting in $n$ different probability 
distributions $p_{i}=p_{\theta}(\cdot | v_i, x, y_{<t})$ with $i=1, 2, \ldots, n$.
Between any two candidate FOVs, we adopt the following distance measure for the discrepancy 
between their decoded token probability distributions
\begin{equation}\label{eqn:JSD_dist}
    d(v_i, v_j) = \mathrm{JSD}(p_{\theta}(\cdot | v_i , x, y_{<t}) \parallel  p_{\theta}(\cdot | v_j,  x, y_{<t}))
\end{equation}
where JSD is the Jensen-Shannon divergence, a symmetric metric that measures the difference 
between two distributions.
With the idea that more different FOV pairs are more likely to include the optimal visual 
context for the current victim token generation, we dynamically select the top $m$ pairs with 
the largest distance according to \eqnref{eqn:JSD_dist}.

\bfsection{Contrastive decoding.}
After obtaining top $m$ visual context pairs with most discrepancies in influencing the token output, we contrast the decoding probability distributions $(p_i, p_j)$ within each pair in order to amplify the information residing in one visual context over the other.
This would potentially recover the victim token over the hallucinated token as the victim token enjoys a sharper contrast in the probability comparisons, especially when one of the visual contexts under comparison is near the optimal grounding.
Specifically, we redistribute the probabilities based on the contrast in log space~\cite{li2022contrastive} for a given FOV pair $(v_i, v_j)$, resulting in the following distribution
\begin{align}\label{eqn:p_fov}
p_{v_{i}/v_{j}} (\cdot | v_i, v_j, x, y_{<t}) &\propto \exp \Big[ (1 + \alpha) f_{\theta} (\cdot | v_i, x, y_{<t}) \nonumber \\
\quad - \alpha & f_{\theta} (\cdot | v_j, x, y_{<t}) \Big]
\end{align}
where $f_{\theta}$ again is the logit distribution, $\alpha$ is the amplification factor 
where larger $\alpha$ indicates a stronger amplification of the differences 
between the distribution pair 
($\alpha = 0$ simplifies \eqnref{eqn:p_fov} to regular decoding from $v_i$ without contrast). 

Unlike existing uni-modal contrastive decoding
methods~\cite{chuang2023dola, gera2023benefits,shi2023trusting} 
that assign an expert and an amateur distribution in the contrast by assuming the final or context-aware layer 
contains more factual knowledge, in our case defining an asymmetric expert distribution among a random pair 
of FOVs is non-trivial. 
For example, the optimal visual context usually resides midway among growing FOVs, making either 
overflowing or insufficient context result in hallucination, as 
seen in \figref{fig:optimal_visual_context}. 
Therefore, as we have no knowledge where the optimal visual context resides, for each pair of FOVs, 
we propose to contrast them bi-directionally, which contains both \textit{positive} 
(larger over smaller-sized FOV) and \textit{negative} (smaller over larger-sized FOV) contrast 
to preserve the completeness of FOV representations (as shown in \figref{fig:halc_overview}).
Essentially, this process results in $2m$ candidate tokens by individual greedy decodings 
which will be further selected by the matching-based beam search algorithm next.

\subsection{Matching-based Beam Search}
\label{subsec:beam_search}

While our adaptive focal-contrast grounding in \secref{subsec:logits_redistribution} focuses on local token corrections at a single generation step, we adopt a sequence-level beam search algorithm~\cite{anderson2016guided} to globally maintain the text generation qualities.
Specifically, with a beam size of $k$, at an HALC decoding step at time $t$, the $k$ beam sequences would generate $2mk$ token candidates for $y_t$ in total from top $m$ focal-contrast pairs.
Different from existing beam score designs~\cite{borgeaud2019leveraging} based only on textual information, we rely on a global visual matching score to select the top $k$ beams from $2mk$ candidates, by comparing the similarity between the current text sequence $y_{\leq t}$ and the original image $v$.
This maintains a diverse but faithful set of generations within the search.
In practice, we employ the Bootstrapping Language-Image Pre-training (BLIP) model~\cite{li2022blip}
for both text and image encoding and compute their similarity scores.
%

%
Combining all components, the full procedure of HALC is summarized in Algorithm~\ref{algo:halc_decoding}.
Notice that by utilizing the fine-grained visual information at different levels for a 
single generation step, we admittedly trade in some computation time for correcting token 
hallucinations. 
The detailed analysis on time complexity is in \appref{app:time_cost}.
One way to increase the HALC decoding speed is through parallelization of decoding from different visual contexts, where we can hope to spend at worst roughly twice of the regular decoding time at HALC steps considering the whole sequence.\footnote{As HALC does not happen at every decoding step. There are also other overhead such as visual grounding affecting the runtime.}

\setlength{\textfloatsep}{8pt}
\begin{algorithm}[t]
\caption{HALC Decoding}
\begin{algorithmic}[1]
    \REQUIRE LVLM {\lvlm}, text query $x$, image input $v$, 
    grounding detector $\mathcal{G}_{d}$,
    FOV sample size $n$, beam size $k$, number of contrast FOV pairs $m$. 
    \OUTPUT Model response $y_{\text{new}}$.
    \REPEAT
        \STATE At every decoding step $t$:
        \FOR{$b = 1$ to beam size $k$}
        \STATE {\lvlm} decoding, obtain current token $y_t^b$
        \IF{$y_t^b \in \{\text{existence}, \text{attribute}, \text{relationship}\}$ \COMMENT{\secref{subsec:token_identify}}}
        \STATE Retrieve visual context $v_d^b \leftarrow \mathcal{G}_{d}(y_t^b, v)$ \COMMENT{\secref{subsec:visual_retrieval}}
        \ENDIF
        \IF{$v_d^b \neq \{\varnothing\}$}
        \STATE Sample $n$ FOVs $v_1, \ldots, v_n$ by expanding $v_d^b$ 
        \ELSE
        \STATE Randomly sample $n$ FOVs $v_1, \ldots, v_n$ from $v$ 
        \ENDIF \COMMENT{\secref{subsec:logits_redistribution}}
        \STATE Compute pair-wise JSDs $d(v_i, v_j), \forall i\neq j$
        \COMMENT{\secref{subsec:logits_redistribution}, \eqnref{eqn:JSD_dist}}
        \STATE Select top-$m$ candidate pairs
        \COMMENT{\secref{subsec:logits_redistribution}}
        \FOR{$i = 1$ to $m$}
        \STATE Apply bi-directional contrast $(p_{{v}_i/{v}_j},p_{{v}_j/{v}_i})$, 
        \STATE get a pair of redistributed logits
        \COMMENT{\secref{subsec:logits_redistribution}, \eqnref{eqn:p_fov}}
        \ENDFOR
        \COMMENT{$y_{\text{new}}^b$ with $2m$ candidates obtained}
        \ENDFOR 
        \STATE Select top $k$ candidates by visual matching
        \COMMENT{\secref{subsec:beam_search}}
        \IF{$v_d^b \neq \{\varnothing\}$ \AND $y_{\text{new}}^b=y_t^b$}
        \STATE $y_{\text{new}}^b$ $\leftarrow$ [IDK]
        \COMMENT{$y_t^b$ is hallucinating, but no correction token was found}
        \ENDIF
        \STATE $y_t^b \leftarrow y_{\text{new}}^b$
        \COMMENT{Hallucinating token $y_t^b$ corrected}
    \UNTIL{each beam has terminated}
\end{algorithmic}
\label{algo:halc_decoding}
\end{algorithm}

\section{Theoretical Analysis on FOV Sampling}

Based on our observation (in \figref{fig:pattern_analysis} and \figref{fig:optimal_visual_context}) that there exists some underlying optimal visual context $v^*$ within the original image $v$ that can largely reduce the object hallucination at the token level, our method aims to recover this optimal visual context $v^*$ based on a sampling process conditioned on $v_d$. 
To do so, we first select the visual contexts, or FOVs, by taking a sequence of FOV samples starting from the initial $v_d$ based on an off-the-shelf detector.
%
While we cannot guarantee that the initial visual grounding $v_d$ is sufficiently accurate to approximate $v^*$ (and directly using $v_d $ could result in unstable behaviors), we could effectively certify the robustness of our FOV sampling strategy in Theorem~\ref{theorem:robustness}. To preserve generality, consider the sampled FOVs are taken from a distribution $\pi(\cdot|v_d)$, 
where $\pi$ can either follow normal distribution sampling around $v_d$, or obey an exponential expansion sampling strategy starting from $v_d$.

\begin{theorem}\label{theorem:robustness}
Let $v^*=(w^*, h^*, p^*)$ be the optimal visual context. Assume there exists a tolerable neighborhood $\mathcal{B}(v^*, \epsilon)=\{\hat{v}: \|\hat{v} - v^*\|\leq \epsilon\}$ around $v^*$, such that decoding from visual contexts within the neighborhood is robust:
\begin{equation}
D(p_\theta(\cdot|v^*), p_\theta(\cdot|\hat{v})) \leq \delta\ll 1,\, \forall \hat{v} \in \mathcal{B}(v^*, \epsilon)
\end{equation}
where $D(\cdot,\cdot)\in [0, 1]$ is a symmetric discrepancy measure between two probability distributions, such as the Jensen-Shannon divergence, or the total variation distance.

Let $v_d=(w_d, h_d, p_d)$ be the initial detection and $v_d=v^*+\eta$ with perturbation $\eta$. The minimum deviation of token probabilities from the optimum with $n$ samples $v_1, v_2, \ldots, v_n$ distributed according to $\pi(\cdot|v_d)$ is denoted as
\begin{equation}\label{eq:min_devi}
    h_{\pi}(v^*, n) = \min_{i=1,\ldots, n} D\left(p_{\theta}(\cdot | v^*), p_{\theta}(\cdot | v_i)\right)
\end{equation}
(a)
For normal distribution sampling $\pi_g(\cdot|v_d)\sim\mathcal{N}(v_d, \sigma^2 I)$, the minimum deviation above is bounded as
\begin{equation}
    h_{\pi_g}(v^*, n)\leq \delta + (1-C_g(\epsilon, \eta; \sigma))^n 
\end{equation}
where $C_g(\epsilon, \eta;\sigma)\in(0, 1)$ is a constant depending on $\epsilon, \eta, \sigma$, and the upper bound goes to $\delta$ when $n\rightarrow\infty$.

(b)
For exponential expansion sampling $\pi_e(\cdot|v_d)\sim\mathcal{U}(r\in[r_{\min}, r_{\max}])$ with samples $v_r=((1+\lambda)^r w_d, (1+\lambda)^r h_d, p_d)$ uniformly from the $r$-space, under the conditions (i) $|p_d-p^*|<\epsilon$ and (ii) $w_d/h_d=w^*/h^*$, the minimum deviation in Eq.~\eqref{eq:min_devi} is bounded below
\begin{equation}
    h_{\pi_e}(v^*, n)\leq \delta + (1-C_e(\epsilon, v^*, v_d;\lambda))^n 
\end{equation}
where $C_e(\epsilon, v^*, v_d;\lambda)\in(0, 1]$ is a constant depending on $\epsilon, v^*, v_d, \lambda$, and the upper bound goes to $\delta$ when $n\rightarrow\infty$.
\end{theorem}
The proof of Theorem~\ref{theorem:robustness} is detailed in \appref{app:robustness}.
The neighborhood radius $\epsilon$ around the optimal $v^*$ can be roughly interpreted as a valid range of optimal visual context to yield 
the correct prediction (e.g., $[v_1, v_2$] in \figref{fig:optimal_visual_context}).
Typically the detection perturbation $\|\eta\|>\epsilon$, making $v_d$ outside of the $\epsilon$-neighborhood of $v^*$.
Through FOV sampling according to some $\pi(\cdot|v_d)$, the above theorem establishes a formal guarantee that at least one of the $n$ samples achieves good approximation of the optimal $v^*$ in the decoding probability space, as the deviation is closer to $\delta$ when $n$ grows.  
The normal sampling distribution, concentrated around $v_d$, is preferred when $v_d$ has minimal perturbations from $v^*$.
And an exponential expansion sampling distribution, with a more averaged coverage of the sampling space, is preferable when less prior of the task is available.
%
%
In practice of our algorithm, we take discrete integer values of $r$ under the exponential expansion distribution for deterministic sampling with $n=4$, acquiring good efficiency and performance.


\section{Experiments}\label{sec:experiments}
\begin{table*}[h!]
\centering
\caption{
    CHAIR evaluation results on MSCOCO dataset of LVLMs with different decoding 
    baselines and SOTAs designed for mitigating OH. 
    Lower $\text{CHAIR}_S$ and $\text{CHAIR}_I$ indicate less OH. 
    Higher BLEU generally represent higher captioning quality, although existing 
    work has reported weak correlation between CHAIR and text overlapping quality metrics.
    Bold indicates the best results of all methods.
}
\setlength{\tabcolsep}{2pt}
\renewcommand{\arraystretch}{0.9}
\resizebox{\linewidth}{!}{
\begin{tabular}{ l | c c c | c c c | c c c}
\toprule
\multirow{2}{*}{Method} 
&\multicolumn{3}{c|}{MiniGPT-4}
&\multicolumn{3}{c|}{LLaVA-1.5}
&\multicolumn{3}{c}{mPLUG-Owl2} \\
\cline{2-10}
&$\text{CHAIR}_S\downarrow$ &$\text{CHAIR}_I\downarrow$ &BLEU$\uparrow$ 
&$\text{CHAIR}_S\downarrow$ &$\text{CHAIR}_I\downarrow$ &BLEU$\uparrow$ 
&$\text{CHAIR}_S\downarrow$ &$\text{CHAIR}_I\downarrow$ &BLEU$\uparrow$   \\ 
\midrule
Greedy 
&$\text{30.87}_{\pm \text{5.45}}$ &$\text{12.33}_{\pm \text{2.07}}$ &$\text{14.33}_{\pm \text{0.00}}$  
&$\text{20.80}_{\pm \text{0.08}}$ &$\text{6.77}_{\pm \text{0.07}}$ &$\text{15.93}_{\pm \text{0.00}}$ 
&$\text{23.20}_{\pm \text{0.35}}$ &$\text{8.33}_{\pm \text{0.28}}$ &$\text{15.37}_{\pm \text{0.00}}$  \\
Beam Search &$\text{29.56}_{\pm \text{6.09}}$ &$\text{11.36}_{\pm \text{0.99}}$ &$\text{14.94}_{\pm \text{0.00}}$ 
&$\text{18.67}_{\pm \text{0.38}}$ &$\text{6.30}_{\pm \text{0.05}}$ &$\text{16.17}_{\pm \text{0.00}}$ 
&$\text{21.67}_{\pm \text{1.61}}$ &$\text{7.63}_{\pm \text{0.40}}$ &$\text{15.77}_{\pm \text{0.00}}$  \\
DoLA &$\text{30.87}_{\pm \text{2.52}}$ &$\text{11.70}_{\pm \text{0.13}}$ &$\text{14.93}_{\pm \text{0.00}}$ 
&$\text{21.00}_{\pm \text{0.67}}$ &$\text{6.70}_{\pm \text{0.38}}$ &$\text{15.93}_{\pm \text{0.00}}$ 
&$\text{24.60}_{\pm \text{0.24}}$ &$\text{8.73}_{\pm \text{0.30}}$ &$\text{15.40}_{\pm \text{0.00}}$\\ 
OPERA 
&$\text{30.00}_{\pm \text{0.43}}$ &$\text{11.67}_{\pm \text{0.22}}$ &$\text{14.87}_{\pm \text{0.00}}$  
&$\text{21.13}_{\pm \text{0.12}}$ &$\text{6.73}_{\pm \text{0.18}}$ &$\text{16.27}_{\pm \text{0.01}}$ 
&$\text{22.13}_{\pm \text{0.86}}$ &$\text{7.57}_{\pm \text{0.16}}$ &$\text{15.53}_{\pm \text{0.00}}$ \\ 
VCD 
&$\text{30.27}_{\pm \text{0.44}}$  &$\text{12.60}_{\pm \text{0.45}}$ &$\text{14.33}_{\pm \text{0.00}}$   
&$\text{23.33}_{\pm \text{5.66}}$  &$\text{7.90}_{\pm \text{0.53}}$ &$\text{14.67}_{\pm \text{0.01}}$ 
&$\text{27.27}_{\pm \text{7.32}}$ &$\text{9.73}_{\pm \text{1.22}}$ &$\text{14.40}_{\pm \text{0.00}}$   \\
Woodpecker &$\text{28.87}_{\pm \text{2.20}}$ 
&$\text{10.20}_{\pm \text{0.85}}$ 
&$\textbf{15.30}_{\pm \text{0.01}}$ 
&$\text{23.85}_{\pm \text{4.62}}$ 
&$\text{7.50}_{\pm \text{0.01}}$ 
&$\textbf{17.05}_{\pm \text{0.00}}$ 
&$\text{26.33}_{\pm \text{1.98}}$
&$\text{8.43}_{\pm \text{0.80}}$ &$\textbf{16.43}_{\pm \text{0.00}}$  \\ 
LURE &$\text{27.88}_{\pm \text{2.25}}$ &$\text{10.20}_{\pm \text{0.85}}$ &$\text{15.03}_{\pm \text{0.11}}$ 
&$\text{19.48}_{\pm \text{2.35}}$ &$\text{6.5}_{\pm \text{0.38}}$ &$\text{15.97}_{\pm \text{0.01}}$   
&$\text{21.27}_{\pm \text{0.06}}$ &$\text{7.67}_{\pm \text{0.16}}$ &$\text{15.65}_{\pm \text{0.05}}$ \\ 
\midrule
\textbf{HALC} &$\textbf{17.80}_{\pm 0.03}$ &$\textbf{8.10}_{\pm \text{0.14}}$ &$\text{14.91}_{\pm \text{0.00}}$ 
&$\textbf{13.80}_{\pm \text{0.08}}$ &$\textbf{5.50}_{\pm \text{0.14}}$ &$\text{16.10}_{\pm \text{0.01}}$ 
&$\textbf{17.33}_{\pm \text{4.30}}$ 
&$\textbf{7.43}_{\pm \text{0.11}}$ 
&$\text{16.27}_{\pm \text{0.00}}$ 
 \\
\bottomrule
\end{tabular}
}
\label{tab:chair_results}
\vspace{-0.1in}
\end{table*}
\bfsection{Benchmarks.} We evaluate HALC on three benchmarks including (1) quantitative metrics CHAIR~\cite{rohrbach2018object} 
and POPE~\cite{li2023evaluating} on MSCOCO~\cite{lin2014microsoft} dataset; 
(2) general-purposed Multimodal Large Language Model Evaluation (MME)~\cite{fu2023mme} benchmark; 
and (3) qualitative evaluation benchmark LLaVA-Bench~\cite{liu2023improved}.
These experiments comprehensively assess HALC's capability on reducing OH in image captioning,
visual-question answering (VQA) and more challenging tasks that generalize to novel domains.
%

\bfsection{Baselines.} To effectively evaluate HALC, besides regular greedy decoding and beam search baselines, we further involve layer-wise contrastive decoding SOTA DoLa~\cite{chuang2023dola}, as well as 
SOTA methods specifically designed to mitigate OH, including 
OPERA~\cite{huang2023opera}, VCD~\cite{leng2023mitigating}, 
Woodpecker~\cite{yin2023woodpecker} 
and LURE~\cite{zhou2023analyzing} in our analysis. All the results are acquired and benchmarked consistently with our unified implementation.
Please refer to Appendix~\ref{app:hyper} for the detailed setting of our HALC.
%
%
%

\bfsection{LVLM Backbones.} Three LVLMs including MiniGPT-4 V2~\cite{chen2023minigpt}, LLaVA-1.5~\cite{liu2023visual} 
and mPLUG-Owl2~\cite{ye2023mplug} are used for both HALC and all aforementioned 
baselines except Woodpecker and LURE, where Woodpecker utilizes 
ChatGPT~\cite{brown2020language} during its self-correction process and LURE distills an extra reviser model from GPT-4~\cite{achiam2023gpt}.

%
%

\subsection{CHAIR and POPE on MSCOCO}\label{subsec:mscoco}
Following existing evaluation 
procedures~\cite{huang2023opera, yin2023woodpecker, liu2023visual},
we randomly sampled 500 images from the validation split of MSCOCO~\cite{lin2014microsoft} 
and conduct evaluations with both CHAIR and POPE.
For each metric, we repeat the experiments five times with different random seeds and report 
average and standard deviations of all the runs.

\bfsection{CHAIR.}
Caption Hallucination Assessment with Image Relevance (CHAIR)~\cite{rohrbach2018object} 
is a tailored tool created to evaluate the occurrence of OH in the task of 
image captioning. 
Specifically, CHAIR measures the extent of OH in an image description by determining 
the proportion of the mentioned objects that are absent in the actual label set. 
This metric includes two separate evaluation aspects: $\text{CHAIR}_S$, which performs 
assessments at the \textit{sentence} level (proportion of the hallucinated sentences over all sentences
), and 
$\text{CHAIR}_I$, which operates at the object \textit{instance} level (proportion of the hallucinated objects over all generated objects).
Lower scores indicate less OH.

We prompt all methods with ``\textit{Please describe this image in detail.}'' and the results 
are illustrated in \tabref{tab:chair_results}.
Besides $\text{CHAIR}_S$ and $\text{CHAIR}_I$, we also report BLEU~\cite{papineni2002bleu}
as an assessment of the text generation quality. 
%
Table~\ref{tab:chair_results} demonstrats that our proposed HALC consistently outperforms
all the existing methods by a large margin.
Notably, a major advantage of HALC is its strong robustness, as can be observed by its much
lower standard deviations, especially when compared to the non-OH specific baselines.
While Woodpecker~\cite{yin2023woodpecker} has the highest generation quality BLEU scores,
this can be largely attributed to the fact that Woodpecker adopts ChatGPT, a much more 
capable LLM, to organize the final outputs, which is not exactly a fair comparison to the 
other methods.

\begin{figure}[t]
    \centering
    \includegraphics[width=.9\linewidth]{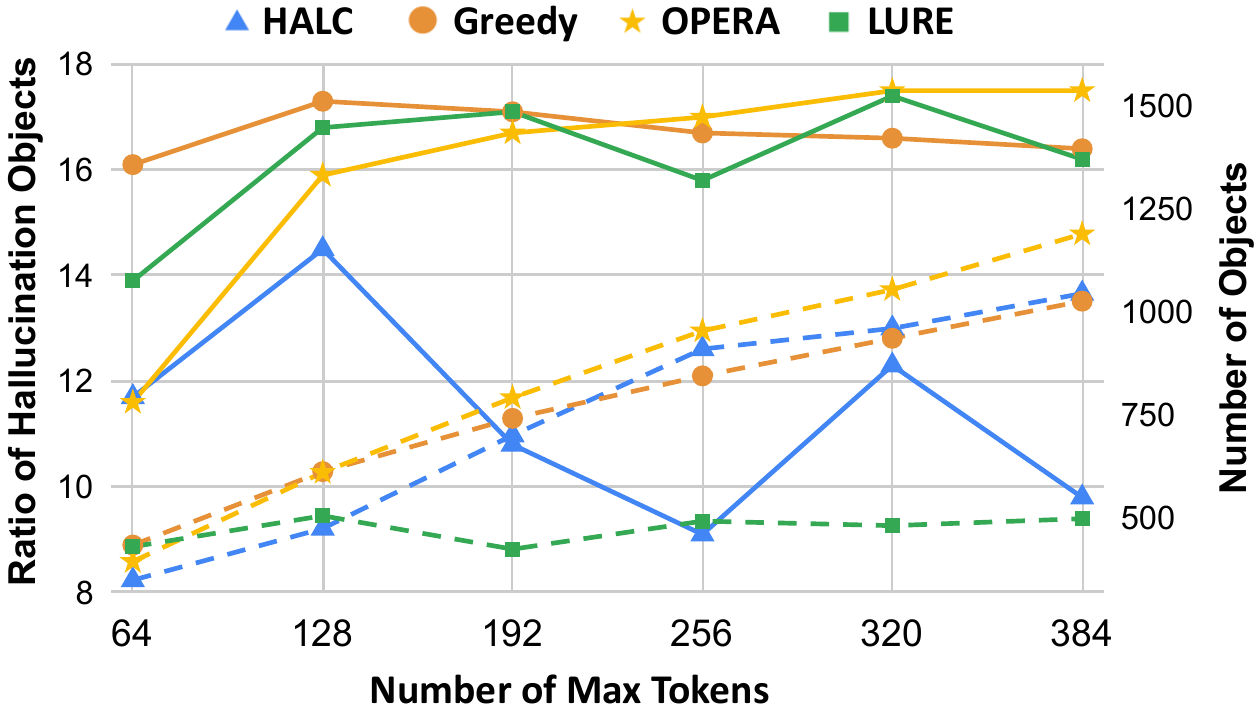}
    \caption{
    Comparing four mainstream methods on the ratio of hallucination objects ($\text{CHAIR}_I$) v.s. the number of max tokens. The right axis (dashed line) indicates the total number of generated objects. HALC outperforms all other methods by maintaining a low ratio of hallucination with the increasing of generated objects. 
    }
    \label{fig:hallu_vs_length}
\end{figure}

We also investigate how HALC performs with longer responses, as showed in 
\figref{fig:hallu_vs_length}, where we plot both the number of generated (dashed) and
hallucinated (solid) objects with randomly sample 100 images.
This experiment is important to further assess HACL's robustness, as it is commonly believed
that OH happens more with objects positioned later in the responses~\cite{zhou2023analyzing}, 
as well as in longer responses~\cite{huang2023opera}.
We observe that HALC is the only method that can keep even smaller number of hallucinations
while the number of generated objects increases, demonstrating its superior performance and
advantageous robustness in reducing OH.

\begin{table*}[ht]
\centering
\caption{
    Proposed OPOPE evaluation results on MSCOCO dataset of LVLMs with different decoding 
    baselines and SOTAs designed for mitigating OH. 
    %
    Higher accuracy, precision, and F score indicate better performance.
    Bold indicates the best results of all methods.
}
\setlength{\tabcolsep}{2pt}
\renewcommand{\arraystretch}{0.9}
\resizebox{\linewidth}{!}{
\begin{tabular}{ l | c c c | c c c | c c c}
\toprule
\multirow{2}{*}{Method} 
&\multicolumn{3}{c|}{MiniGPT-4}
&\multicolumn{3}{c|}{LLaVA-1.5}
&\multicolumn{3}{c}{mPLUG-Owl2} \\
\cline{2-10}
&Accuracy$\uparrow$ &Precision$\uparrow$ &$F_{\beta=0.2}\uparrow$  
&Accuracy$\uparrow$ &Precision$\uparrow$ &$F_{\beta=0.2}\uparrow$  
&Accuracy$\uparrow$ &Precision$\uparrow$ &$F_{\beta=0.2}\uparrow$   \\ 
\midrule
Greedy 
&$\text{66.78}_{\pm \text{1.27}}$ &$\text{90.43}_{\pm \text{25.1}}$ &$\text{85.79}_{\pm \text{18.7}}$  
&$\text{70.56}_{\pm \text{1.51}}$ &$\text{91.08}_{\pm \text{20.6}}$ &$\text{87.72}_{\pm \text{16.3}}$
&$\text{69.77}_{\pm \text{1.18}}$ &$\text{91.07}_{\pm \text{17.8}}$ &$\text{87.45}_{\pm \text{13.9}}$  \\
Beam Search 
&$\text{67.22}_{\pm \text{0.74}}$ &$\text{91.20}_{\pm \text{14.4}}$ &$\text{86.57}_{\pm \text{10.8}}$  
&$\text{69.87}_{\pm \text{1.37}}$ &$\text{91.72}_{\pm \text{20.4}}$ &$\text{88.01}_{\pm \text{15.97}}$  
&$\text{69.20}_{\pm \text{0.90}}$ &$\text{91.90}_{\pm \text{15.1}}$ &$\text{87.91}_{\pm \text{11.7}}$ \\
DoLA 
&$\text{67.06}_{\pm \text{1.19}}$ &$\text{90.84}_{\pm \text{23.1}}$ &$\text{86.22}_{\pm \text{17.3}}$ 
&$\textbf{70.69}_{\pm \text{1.50}}$ &$\text{90.87}_{\pm \text{19.8}}$ &$\text{87.59}_{\pm \text{15.74}}$ 
&$\textbf{70.17}_{\pm \text{1.69}}$ &$\text{91.97}_{\pm \text{24.5}}$ &$\text{88.30}_{\pm \text{19.26}}$ \\ 
OPERA 
&$\text{67.26}_{\pm \text{1.04}}$ &$\text{90.76}_{\pm \text{20.0}}$ &$\text{86.25}_{\pm \text{15.0}}$ 
&$\text{69.73}_{\pm \text{1.34}}$ &$\text{91.10}_{\pm \text{19.4}}$ &$\text{87.46}_{\pm \text{15.3}}$  
&$\text{69.26}_{\pm \text{0.45}}$ &$\textbf{93.06}_{\pm \text{8.01}}$ &$\textbf{88.83}_{\pm \text{6.14}}$ \\ 
VCD 
&$\text{65.78}_{\pm \text{0.96}}$  &$\text{90.02}_{\pm \text{20.7}}$ &$\text{85.00}_{\pm \text{15.1}}$  
&$\text{70.67}_{\pm \text{1.22}}$  &$\text{91.62}_{\pm \text{16.7}}$ &$\text{88.19}_{\pm \text{13.3}}$
&$\text{69.81}_{\pm \text{0.65}}$ &$\text{92.70}_{\pm \text{11.0}}$ &$\text{88.76}_{\pm \text{8.49}}$  \\
Woodpecker 
&$\text{67.78}_{\pm \text{0.88}}$ &$\text{91.33}_{\pm \text{16.66}}$ &$\text{86.91}_{\pm \text{12.6}}$  
&$\text{69.80}_{\pm \text{0.54}}$ &$\text{91.80}_{\pm \text{8.41}}$ &$\text{88.04}_{\pm \text{6.56}}$   
&$\text{68.90}_{\pm \text{1.02}}$ &$\text{92.22}_{\pm \text{17.98}}$ &$\text{88.05}_{\pm \text{13.77}}$   \\ 
LURE 
&$\textbf{68.14}_{\pm \text{0.99}}$ &$\text{90.95}_{\pm \text{17.34}}$ &$\text{86.76}_{\pm \text{13.23}}$  
&$\text{70.00}_{\pm \text{1.53}}$ &$\text{90.89}_{\pm \text{21.9}}$ &$\text{87.38}_{\pm \text{17.3}}$  
&$\text{69.24}_{\pm \text{1.60}}$ &$\text{90.54}_{\pm \text{23.37}}$ &$\text{86.85}_{\pm \text{18.28}}$  \\ 
\midrule
\textbf{HALC} 
&$\text{66.76}_{\pm \text{0.68}}$ &$\textbf{91.95}_{\pm \text{15.0}}$ &$\textbf{86.92}_{\pm \text{11.1}}$  
&$\text{70.59}_{\pm \text{0.82}}$ &$\textbf{92.94}_{\pm \text{12.18}}$ &$\textbf{89.22}_{\pm \text{9.55}}$
&$\text{70.12}_{\pm \text{0.98}}$ &$\text{91.94}_{\pm \text{15.1}}$ &$\text{88.26}_{\pm \text{11.85}}$  \\
\bottomrule
\end{tabular}
}
\label{tab:pope_avg_results}
\vspace{-0.1in}
\end{table*}
\bfsection{POPE.}
Polling-based Object Probing Evaluation (POPE)~\cite{li2023evaluating} evaluates OH via a 
streamlined approach, which incorporates a list of yes-or-no questions to prompt 
LVLMs for presence of positive and negative objects.
%
%
When selecting negative (non-existing) objects for prompting, POPE provides three sampling
options: random, popular, and adversarial.
We refer detailed explanations of the different options to its original 
paper~\cite{li2023evaluating}.
%
%

One distinct difference between POPE and CHAIR is that POPE relies on interacting with the 
examined LVLM directly.
While this requirement is not an issue when evaluating the decoding-based 
baselines, it limits its adaptation to post-hoc methods such as LURE~\cite{zhou2023analyzing}.
It also creates larger instabilities when the examined LVLM incorporates smaller language 
backbones such as LLaMA-7B~\cite{touvron2023llama}, which has less robust chat capability.
To these concerns, we propose \textit{offline POPE (OPOPE)}, which keeps the object
sampling and yes/no query strategy from POPE, but replaces the live interactions with 
offline checks.
Specifically, instead of querying the model with ``\textit{Is there a \{\} in the image?}'', 
where ``\textit{\{\}}'' is the queried object, we first ask the examined LVLM to give its 
detailed descriptions of the image, and then manually check if the sampled positive/negative 
objects exist in the captions when computing the OPOPE scores.

We also adjust the main metrics for comparison. As it is more random for descriptions to include 
the exact sampled hallucinated objects, false-negative (FN) and the resulting recall
become less trustable in the offline checks.
%
Therefore, we propose to use F-beta, instead of F-1, as the main metric of OPOPE, so that the
final score relies less on the FN.
Specifically, we have 
$F_\beta = (1+\beta^2)\cdot(\text{precision}\cdot\text{recall})/(\beta^2\cdot\text{precision}+\text{recall})$, 
where we use $\beta = 0.2$ throughout our experiments.
The evaluation results incorporating OPOPE is shown in \tabref{tab:pope_avg_results}.
All the numbers are averaged results of the three sampling methods (random, popular and 
adversarial, as in the original POPE), while the complete version of the table 
is shown in \appref{app:opope}.
We also include the original POPE evaluation results in \appref{app:pope} where
HALC also outperforms other methods in most of the settings.

\subsection{MME}\label{subsec:mme}
\begin{figure}[t]
    \centering
    \includegraphics[width=0.9\linewidth]{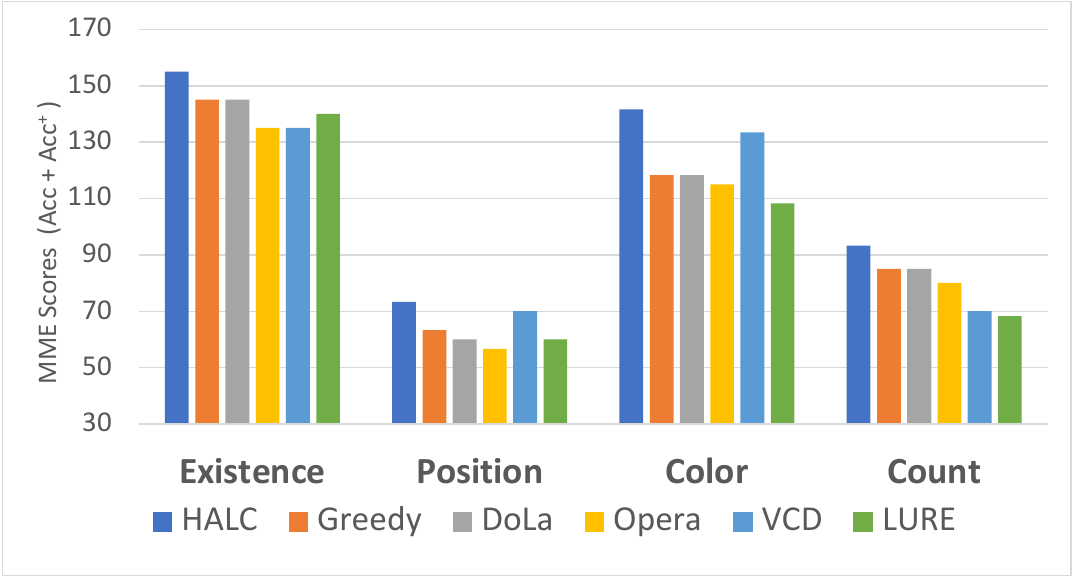}
    \caption{
    Comparison across OH baselines and SOTAs on four OH-critical MME subsets. 
    All methods adopt MiniGPT-4 as LVLM backbone. 
    HALC outperforms all other methods with a large margin: 
    \textit{existence}: +10.7\%; \textit{position}: +18.3\%; \textit{color}: +19.4\% and 
    \textit{count}: +20.2\% in average. 
    }
    \label{fig:mme_result}
\end{figure}

The Multimodal Large Language Model Evaluation (MME)~\cite{fu2023mme} benchmark is a 
comprehensive tool designed to quantitatively compare multimodal LLMs.
%
%
%
Following~\citet{yin2023woodpecker, leng2023mitigating}, we utilize the ``existence" 
and ``count" subsets to evaluate the object existence hallucinations and the
``position" and ``color" subsets for object attribute and relationship hallucination. 
%
%
Please refer to \appref{app:mme} for experiment details. 
The comprehensive results across six methods are reported in \figref{fig:mme_result}, where 
HALC significantly outperforms all the other methods on each sub-task, indicating an overall 
performance gain in reducing OH while preserving generation quality.
%
%
%

\subsection{LLaVA-Bench Qualitative Study}\label{subsec:llava_bench}
LLaVA-Bench~\cite{liu2023improved} is a collection of 24 images, where each image is paired 
with a detailed, manually-crafted description and carefully selected questions.
The questions are divided into three categories: simple QA (conversation), detailed 
descriptions, and complex reasoning.
In this experiment, we leverage LLaVA-Bench as a case study to qualitatively compare the 
decoding outputs of HALC with other methods. 
The results are shown in \appref{app:llava_bench}.

\section{Analysis and Ablation Studies}\label{sec:ablation}

\subsection{Adaptive Focal-contrast Grounding}
\vspace{-0.05in}
\bfsection{FOV Sampling initialization.}
The visual context retrieval process described in \secref{subsec:visual_retrieval} utilizes 
detector output as a key component of the adaptive focal-contrast grounding algorithm introduced in \secref{subsec:logits_redistribution}. 
However, it is important to note that HALC primarily uses the detector output as a 
\textit{initialization} for the field of view (FOV) sampling process, rather than depending 
heavily on it.
In this section, we present empirical results to compare different methods of sampling 
initialization, which include random sampling (selecting a random FOV within the image), 
center initialization (selecting a fixed region in the center of the image), original 
image initialization (using the entire image) and detector initialization (using the 
detector output).
More specifically, we include an extra detector model, OWLv2~\citep{minderer2024scaling}, 
in addition to the Grounding Dino~\citep{liu2023grounding} illustrated in previous sections.
\begin{table}[ht]
\vspace{-0.1in}
\centering
\caption{
    HALC performance with different sampling initialization.
}
\setlength{\tabcolsep}{2pt}
\renewcommand{\arraystretch}{0.9}
\resizebox{\linewidth}{!}{
\begin{tabular}{ l | c c c c c}
\toprule
Init. &$\text{CHAIR}_S\downarrow$ &$\text{CHAIR}_I\downarrow$ &$\text{OPOPE}\uparrow$ 
&$\text{POPE}\uparrow$ &$\text{BLEU}\uparrow$  \\ 
\midrule
Random 
&$\text{25.6}$ &$\text{11.8}$ &$\text{83.33}$  &$\text{67.67}$ &$\text{15.10}$ \\
Center 
&$\text{23.9}$ &$\text{11.2}$ &$\text{86.62}$  &$\text{69.10}$ &$\text{14.80}$ \\
Original 
&$\text{27.8}$ &$\text{12.2}$ &$\text{85.20}$  &$\text{68.33}$ &$\text{15.50}$ \\ 
G. Dino
&$\textbf{22.0}$ &$\textbf{8.8}$ &$\textbf{88.20}$  &$\textbf{70.67}$ &$\textbf{16.40}$ \\ 
OWLv2 
&$\text{23.4}$ &$\text{10.8}$ &$\text{84.47}$  &$\text{67.50}$ &$\text{15.70}$ \\ 
\bottomrule
\end{tabular}
}
\label{tab:initialization_ablation}
\vspace{-0.2in}
\end{table}

As shown in \tabref{tab:initialization_ablation}, both random and center initialization 
perform better than using the original image as the visual input. 
This result confirms the robustness of the proposed FOV sampling process. 
Additionally, both detectors deliver better performance than the other initializations, further demonstrating that using a detector-grounded FOV provides an 
effective starting point for the subsequent conditional FOV sampling process.
%
%

\bfsection{Exponential Expanding ratio.}
Besides initialization, another important parameter used in adaptive focal-contrast
grounding is the expanding ratio $\lambda$, which determines each sampling FOV as in 
\eqnref{eq:fov_expand}.
Thus we further analyze the performance of HALC with different expanding ratios.
\begin{table}[ht]
\vspace{-0.1in}
\centering
\caption{
    HALC performance with different expanding ratios.
}
\setlength{\tabcolsep}{2pt}
\renewcommand{\arraystretch}{0.9}
\resizebox{\linewidth}{!}{
\begin{tabular}{ l | c c c c c}
\toprule
$\lambda$~~~~~&$\text{CHAIR}_S\downarrow$ &$\text{CHAIR}_I\downarrow$ &$\text{OPOPE}\uparrow$ 
&$\text{POPE}\uparrow$ &$\text{BLEU}\uparrow$  \\ 
\midrule
0.2 
&$\text{22.0}$ &$\text{8.5}$ &$\text{86.45}$  &$\text{69.63}$ &$\textbf{16.60}$ \\
0.4 
&$\textbf{18.0}$ &$\textbf{7.6}$ &$\text{87.33}$  &$\text{70.20}$ &$\text{16.10}$ \\
0.6
&$\text{22.0}$ &$\text{8.8}$ &$\textbf{88.20}$  &$\textbf{70.67}$ &$\text{16.40}$ \\
0.8
&$\text{28.0}$ &$\text{9.6}$ &$\text{86.45}$  &$\text{69.63}$ &$\text{14.80}$ \\ 
1.0 
&$\text{26.0}$ &$\text{8.9}$ &$\text{84.32}$  &$\text{69.63}$ &$\text{14.70}$ \\ 
\bottomrule
\end{tabular}
}
\label{tab:expanding_ratios}
\vspace{-0.2in}
\end{table}

\tabref{tab:expanding_ratios} demonstrates that an expanding ratio of 0.6 is optimal. 
We hypothesize that the poorer performance associated with smaller or larger expanding ratios 
is due to that smaller ratios increase the number of FOV samples, which presents greater 
challenges for the global beam search.
On the other hand, larger ratios decrease the granularity of the FOV in the image, potentially
leading to more severe hallucinations.

\subsection{Global Beam Search}\label{subsec:scoring_models}
\vspace{-0.05in}
\bfsection{Beam sizes.}
As is common with all beam search algorithms, beam size $k$ is a major hyperparameter.
Thus here we examine the performance of HALC w.r.t. different values of $k$.
\begin{table}[ht]
\vspace{-0.2in}
\centering
\caption{
    HALC performance with different values of beam size $k$.
}
\setlength{\tabcolsep}{2pt}
\renewcommand{\arraystretch}{0.9}
\resizebox{\linewidth}{!}{
\begin{tabular}{ l | c c c c c}
\toprule
$k$~~~&$\text{CHAIR}_S\downarrow$ &$\text{CHAIR}_I\downarrow$ &$\text{OPOPE}\uparrow$ 
&$\text{POPE}\uparrow$ &$\text{BLEU}\uparrow$  \\ 
\midrule
1 
&$\text{36.0}$ &$\text{14.6}$ &$\text{88.20}$  &$\text{70.49}$ &$\text{15.40}$ \\
2 
&$\textbf{22.0}$ &$\textbf{8.8}$ &$\textbf{88.74}$  &$\textbf{70.67}$ &$\textbf{16.40}$ \\
3 
&$\text{26.0}$ &$\text{9.8}$ &$\text{87.65}$  &$\textbf{70.67}$ &$\text{15.40}$ \\ 
5
&$\text{29.6}$ &$\text{11.1}$ &$\text{86.33}$  &$\text{70.14}$ &$\text{15.70}$ \\ 
8
&$\text{33.3}$ &$\text{13.8}$ &$\text{87.73}$  &$\text{70.14}$ &$\text{15.50}$ \\
\bottomrule
\end{tabular}
}
\label{tab:beam_sizes}
\vspace{-0.2in}
\end{table}

\tabref{tab:beam_sizes} shows improved performance as the beam 
size initially increases from one.
However, when the beam size reaches or exceeds two, the number of FOV samples also increases, 
making it more challenging for the global beam search module to select the optimal visual 
context from all the samples, thus leading to a higher rate of hallucination. 
Furthermore, as the beam size continues to increase, the variance of HALC's performance also increases, indicating that it will be more difficult to select the top candidate as the global matching model also suffers from hallucination.
%

\bfsection{Scoring methods.}
Finally, we compare the BLIP and CLIP scoring models with random selection to rank the beams.
\begin{table}[ht]
\vspace{-0.1in}
\centering
\caption{
    HALC performance with different scoring methods.
}
\setlength{\tabcolsep}{2pt}
\renewcommand{\arraystretch}{0.9}
\resizebox{\linewidth}{!}{
\begin{tabular}{ l | c c c c c}
\toprule
&$\text{CHAIR}_S\downarrow$ &$\text{CHAIR}_I\downarrow$ &$\text{OPOPE}\uparrow$ 
&$\text{POPE}\uparrow$ &$\text{BLEU}\uparrow$  \\ 
\midrule
Random
&$\text{26.6}$ &$\text{12.8}$ &$\text{85.45}$  &$\text{68.45}$ &$\text{15.20}$ \\
BLIP 
&$\textbf{22.0}$ &$\textbf{8.8}$ &$\textbf{88.20}$  &$\text{70.67}$ &$\textbf{16.40}$ \\
CLIP 
&$\text{23.4}$ &$\text{10.0}$ &$\text{87.67}$  &$\textbf{71.96}$ &$\text{15.60}$ \\ 
\bottomrule
\end{tabular}
}
\label{tab:scoring_model}
\vspace{-0.18in}
\end{table}

As shown in \tabref{tab:scoring_model}, different scoring methods do not lead to large 
variations and they all outperform random selection.

\vspace{-0.1in}
\section{Conclusion}\label{sec:conclusion}
We present HALC, a novel decoding algorithm designed to mitigate OH in LVLMs.
HALC operates on both local and global levels, integrating a robust adaptive focal-contrast 
grounding mechanism to better utilize fine-grained visual information for correcting hallucinated tokens, and a specialized beam search algorithm that promotes further visually matched generations.
Comprehensive experiments demonstrate that HALC effectively reduces OH, achieving SOTA
performance while preserving sequence generation quality, and can be conveniently integrated into 
existing LVLMs without additional training or data. 
%
A benchmarking tool was also built to support convenient 
comparisons across all available OH reduction strategies comprehensively. 
\newpage
\section*{Impact Statement}
This paper presents work whose goal is to advance the field of Machine Learning. 
There are many potential societal consequences of our work, none which we feel must be 
specifically highlighted here.
\section*{Acknowledgement}
We thank Lingyu Gao for initial discussion and constructive suggestions.
This work was supported in part by the Research Computing Center at the 
University of Chicago, and Cisco Faculty Research Award.
We also thank Center for AI Safety and Google Cloud Research Credits program 
for supporting our computing needs.
Any opinions, findings, conclusions, or recommendations expressed in this 
material are those of the authors and do not necessarily reflect the views of
any funding agencies. 
\bibliography{references}

\begin{thebibliography}{46}
\providecommand{\natexlab}[1]{#1}
\providecommand{\url}[1]{\texttt{#1}}
\expandafter\ifx\csname urlstyle\endcsname\relax
  \providecommand{\doi}[1]{doi: #1}\else
  \providecommand{\doi}{doi: \begingroup \urlstyle{rm}\Url}\fi

\bibitem[Achiam et~al.(2023)Achiam, Adler, Agarwal, Ahmad, Akkaya, Aleman, Almeida, Altenschmidt, Altman, Anadkat, et~al.]{achiam2023gpt}
Achiam, J., Adler, S., Agarwal, S., Ahmad, L., Akkaya, I., Aleman, F.~L., Almeida, D., Altenschmidt, J., Altman, S., Anadkat, S., et~al.
\newblock Gpt-4 technical report.
\newblock \emph{arXiv preprint arXiv:2303.08774}, 2023.

\bibitem[Anderson et~al.(2016)Anderson, Fernando, Johnson, and Gould]{anderson2016guided}
Anderson, P., Fernando, B., Johnson, M., and Gould, S.
\newblock Guided open vocabulary image captioning with constrained beam search.
\newblock \emph{arXiv preprint arXiv:1612.00576}, 2016.

\bibitem[Biber et~al.(2000)Biber, Johansson, Leech, Conrad, and Finegan]{biber2000longman}
Biber, D., Johansson, S., Leech, G., Conrad, S., and Finegan, E.
\newblock Longman grammar of spoken and written english, 2000.

\bibitem[Biten et~al.(2022)Biten, G{\'o}mez, and Karatzas]{biten2022let}
Biten, A.~F., G{\'o}mez, L., and Karatzas, D.
\newblock Let there be a clock on the beach: Reducing object hallucination in image captioning.
\newblock In \emph{Proceedings of the IEEE/CVF Winter Conference on Applications of Computer Vision}, pp.\  1381--1390, 2022.

\bibitem[Borgeaud \& Emerson(2019)Borgeaud and Emerson]{borgeaud2019leveraging}
Borgeaud, S. and Emerson, G.
\newblock Leveraging sentence similarity in natural language generation: Improving beam search using range voting.
\newblock \emph{arXiv preprint arXiv:1908.06288}, 2019.

\bibitem[Brown et~al.(2020)Brown, Mann, Ryder, Subbiah, Kaplan, Dhariwal, Neelakantan, Shyam, Sastry, Askell, et~al.]{brown2020language}
Brown, T., Mann, B., Ryder, N., Subbiah, M., Kaplan, J.~D., Dhariwal, P., Neelakantan, A., Shyam, P., Sastry, G., Askell, A., et~al.
\newblock Language models are few-shot learners.
\newblock \emph{Advances in neural information processing systems}, 33:\penalty0 1877--1901, 2020.

\bibitem[Chen et~al.(2023)Chen, Zhu, Shen, Li, Liu, Zhang, Krishnamoorthi, Chandra, Xiong, and Elhoseiny]{chen2023minigpt}
Chen, J., Zhu, D., Shen, X., Li, X., Liu, Z., Zhang, P., Krishnamoorthi, R., Chandra, V., Xiong, Y., and Elhoseiny, M.
\newblock Minigpt-v2: large language model as a unified interface for vision-language multi-task learning.
\newblock \emph{arXiv preprint arXiv:2310.09478}, 2023.

\bibitem[Chuang et~al.(2023)Chuang, Xie, Luo, Kim, Glass, and He]{chuang2023dola}
Chuang, Y.-S., Xie, Y., Luo, H., Kim, Y., Glass, J., and He, P.
\newblock Dola: Decoding by contrasting layers improves factuality in large language models.
\newblock \emph{arXiv preprint arXiv:2309.03883}, 2023.

\bibitem[Cui et~al.(2023)Cui, Zhou, Yang, Wu, Zhang, Zou, and Yao]{cui2023holistic}
Cui, C., Zhou, Y., Yang, X., Wu, S., Zhang, L., Zou, J., and Yao, H.
\newblock Holistic analysis of hallucination in gpt-4v (ision): Bias and interference challenges.
\newblock \emph{arXiv preprint arXiv:2311.03287}, 2023.

\bibitem[Dai et~al.(2022)Dai, Liu, Ji, Su, and Fung]{dai2022plausible}
Dai, W., Liu, Z., Ji, Z., Su, D., and Fung, P.
\newblock Plausible may not be faithful: Probing object hallucination in vision-language pre-training.
\newblock \emph{arXiv preprint arXiv:2210.07688}, 2022.

\bibitem[Dai et~al.(2023)Dai, Li, Li, Tiong, Zhao, Wang, Li, Fung, and Hoi]{Dai2023InstructBLIPTG}
Dai, W., Li, J., Li, D., Tiong, A. M.~H., Zhao, J., Wang, W., Li, B.~A., Fung, P., and Hoi, S. C.~H.
\newblock Instructblip: Towards general-purpose vision-language models with instruction tuning.
\newblock \emph{ArXiv}, abs/2305.06500, 2023.
\newblock URL \url{https://api.semanticscholar.org/CorpusID:258615266}.

\bibitem[Daunhawer et~al.(2021)Daunhawer, Sutter, Chin-Cheong, Palumbo, and Vogt]{daunhawer2021limitations}
Daunhawer, I., Sutter, T.~M., Chin-Cheong, K., Palumbo, E., and Vogt, J.~E.
\newblock On the limitations of multimodal vaes.
\newblock \emph{arXiv preprint arXiv:2110.04121}, 2021.

\bibitem[Fu et~al.(2023)Fu, Chen, Shen, Qin, Zhang, Lin, Yang, Zheng, Li, Sun, et~al.]{fu2023mme}
Fu, C., Chen, P., Shen, Y., Qin, Y., Zhang, M., Lin, X., Yang, J., Zheng, X., Li, K., Sun, X., et~al.
\newblock Mme: A comprehensive evaluation benchmark for multimodal large language models.
\newblock \emph{arXiv preprint arXiv:2306.13394}, 2023.

\bibitem[Gera et~al.(2023)Gera, Friedman, Arviv, Gunasekara, Sznajder, Slonim, and Shnarch]{gera2023benefits}
Gera, A., Friedman, R., Arviv, O., Gunasekara, C., Sznajder, B., Slonim, N., and Shnarch, E.
\newblock The benefits of bad advice: Autocontrastive decoding across model layers.
\newblock \emph{arXiv preprint arXiv:2305.01628}, 2023.

\bibitem[Guan et~al.(2023)Guan, Liu, Li, Wang, Yacoob, and Zhou]{guan2023hallusionbench}
Guan, T., Liu, F., Li, X. W. R. X.~Z., Wang, X. L.~X., Yacoob, L. C. F. H.~Y., and Zhou, D. M.~T.
\newblock Hallusionbench: An advanced diagnostic suite for entangled language hallucination \& visual illusion in large vision-language models.
\newblock \emph{arXiv e-prints}, pp.\  arXiv--2310, 2023.

\bibitem[Gunjal et~al.(2023)Gunjal, Yin, and Bas]{gunjal2023detecting}
Gunjal, A., Yin, J., and Bas, E.
\newblock Detecting and preventing hallucinations in large vision language models.
\newblock \emph{arXiv preprint arXiv:2308.06394}, 2023.

\bibitem[Honnibal \& Montani(2017)Honnibal and Montani]{honnibal2017spacy}
Honnibal, M. and Montani, I.
\newblock spacy 2: Natural language understanding with bloom embeddings, convolutional neural networks and incremental parsing.
\newblock \emph{To appear}, 7\penalty0 (1):\penalty0 411--420, 2017.

\bibitem[Huang et~al.(2023)Huang, Dong, Zhang, Wang, He, Wang, Lin, Zhang, and Yu]{huang2023opera}
Huang, Q., Dong, X., Zhang, P., Wang, B., He, C., Wang, J., Lin, D., Zhang, W., and Yu, N.
\newblock Opera: Alleviating hallucination in multi-modal large language models via over-trust penalty and retrospection-allocation.
\newblock \emph{arXiv preprint arXiv:2311.17911}, 2023.

\bibitem[Leng et~al.(2023)Leng, Zhang, Chen, Li, Lu, Miao, and Bing]{leng2023mitigating}
Leng, S., Zhang, H., Chen, G., Li, X., Lu, S., Miao, C., and Bing, L.
\newblock Mitigating object hallucinations in large vision-language models through visual contrastive decoding.
\newblock \emph{arXiv preprint arXiv:2311.16922}, 2023.

\bibitem[Li et~al.(2022{\natexlab{a}})Li, Li, Xiong, and Hoi]{li2022blip}
Li, J., Li, D., Xiong, C., and Hoi, S.
\newblock Blip: Bootstrapping language-image pre-training for unified vision-language understanding and generation.
\newblock In \emph{International Conference on Machine Learning}, pp.\  12888--12900. PMLR, 2022{\natexlab{a}}.

\bibitem[Li et~al.(2019)Li, Yatskar, Yin, Hsieh, and Chang]{li2019visualbert}
Li, L.~H., Yatskar, M., Yin, D., Hsieh, C.-J., and Chang, K.-W.
\newblock Visualbert: A simple and performant baseline for vision and language.
\newblock \emph{arXiv preprint arXiv:1908.03557}, 2019.

\bibitem[Li et~al.(2022{\natexlab{b}})Li, Holtzman, Fried, Liang, Eisner, Hashimoto, Zettlemoyer, and Lewis]{li2022contrastive}
Li, X.~L., Holtzman, A., Fried, D., Liang, P., Eisner, J., Hashimoto, T., Zettlemoyer, L., and Lewis, M.
\newblock Contrastive decoding: Open-ended text generation as optimization.
\newblock \emph{arXiv preprint arXiv:2210.15097}, 2022{\natexlab{b}}.

\bibitem[Li et~al.(2023)Li, Du, Zhou, Wang, Zhao, and Wen]{li2023evaluating}
Li, Y., Du, Y., Zhou, K., Wang, J., Zhao, W.~X., and Wen, J.-R.
\newblock Evaluating object hallucination in large vision-language models.
\newblock \emph{arXiv preprint arXiv:2305.10355}, 2023.

\bibitem[Lin et~al.(2014)Lin, Maire, Belongie, Hays, Perona, Ramanan, Doll{\'a}r, and Zitnick]{lin2014microsoft}
Lin, T.-Y., Maire, M., Belongie, S., Hays, J., Perona, P., Ramanan, D., Doll{\'a}r, P., and Zitnick, C.~L.
\newblock Microsoft coco: Common objects in context.
\newblock In \emph{Computer Vision--ECCV 2014: 13th European Conference, Zurich, Switzerland, September 6-12, 2014, Proceedings, Part V 13}, pp.\  740--755. Springer, 2014.

\bibitem[Liu et~al.(2023{\natexlab{a}})Liu, Li, Li, and Lee]{liu2023improved}
Liu, H., Li, C., Li, Y., and Lee, Y.~J.
\newblock Improved baselines with visual instruction tuning.
\newblock \emph{arXiv preprint arXiv:2310.03744}, 2023{\natexlab{a}}.

\bibitem[Liu et~al.(2023{\natexlab{b}})Liu, Li, Wu, and Lee]{liu2023visual}
Liu, H., Li, C., Wu, Q., and Lee, Y.~J.
\newblock Visual instruction tuning.
\newblock \emph{arXiv preprint arXiv:2304.08485}, 2023{\natexlab{b}}.

\bibitem[Liu et~al.(2023{\natexlab{c}})Liu, Zeng, Ren, Li, Zhang, Yang, Li, Yang, Su, Zhu, et~al.]{liu2023grounding}
Liu, S., Zeng, Z., Ren, T., Li, F., Zhang, H., Yang, J., Li, C., Yang, J., Su, H., Zhu, J., et~al.
\newblock Grounding dino: Marrying dino with grounded pre-training for open-set object detection.
\newblock \emph{arXiv preprint arXiv:2303.05499}, 2023{\natexlab{c}}.

\bibitem[Minderer et~al.(2023)Minderer, Gritsenko, and Houlsby]{minderer2023scaling}
Minderer, M., Gritsenko, A., and Houlsby, N.
\newblock Scaling open-vocabulary object detection.
\newblock \emph{arXiv preprint arXiv:2306.09683}, 2023.

\bibitem[Minderer et~al.(2024)Minderer, Gritsenko, and Houlsby]{minderer2024scaling}
Minderer, M., Gritsenko, A., and Houlsby, N.
\newblock Scaling open-vocabulary object detection.
\newblock \emph{Advances in Neural Information Processing Systems}, 36, 2024.

\bibitem[Papineni et~al.(2002)Papineni, Roukos, Ward, and Zhu]{papineni2002bleu}
Papineni, K., Roukos, S., Ward, T., and Zhu, W.-J.
\newblock Bleu: a method for automatic evaluation of machine translation.
\newblock In \emph{Proceedings of the 40th annual meeting of the Association for Computational Linguistics}, pp.\  311--318, 2002.

\bibitem[Radford et~al.(2021)Radford, Kim, Hallacy, Ramesh, Goh, Agarwal, Sastry, Askell, Mishkin, Clark, et~al.]{radford2021learning}
Radford, A., Kim, J.~W., Hallacy, C., Ramesh, A., Goh, G., Agarwal, S., Sastry, G., Askell, A., Mishkin, P., Clark, J., et~al.
\newblock Learning transferable visual models from natural language supervision.
\newblock In \emph{International conference on machine learning}, pp.\  8748--8763. PMLR, 2021.

\bibitem[Rohrbach et~al.(2018)Rohrbach, Hendricks, Burns, Darrell, and Saenko]{rohrbach2018object}
Rohrbach, A., Hendricks, L.~A., Burns, K., Darrell, T., and Saenko, K.
\newblock Object hallucination in image captioning.
\newblock \emph{arXiv preprint arXiv:1809.02156}, 2018.

\bibitem[Shi et~al.(2023)Shi, Han, Lewis, Tsvetkov, Zettlemoyer, and Yih]{shi2023trusting}
Shi, W., Han, X., Lewis, M., Tsvetkov, Y., Zettlemoyer, L., and Yih, S. W.-t.
\newblock Trusting your evidence: Hallucinate less with context-aware decoding.
\newblock \emph{arXiv preprint arXiv:2305.14739}, 2023.

\bibitem[Touvron et~al.(2023)Touvron, Lavril, Izacard, Martinet, Lachaux, Lacroix, Rozi{\`e}re, Goyal, Hambro, Azhar, et~al.]{touvron2023llama}
Touvron, H., Lavril, T., Izacard, G., Martinet, X., Lachaux, M.-A., Lacroix, T., Rozi{\`e}re, B., Goyal, N., Hambro, E., Azhar, F., et~al.
\newblock Llama: Open and efficient foundation language models.
\newblock \emph{arXiv preprint arXiv:2302.13971}, 2023.

\bibitem[Tu et~al.(2023)Tu, Cui, Wang, Zhou, Zhao, Han, Zhou, Yao, and Xie]{tu2023many}
Tu, H., Cui, C., Wang, Z., Zhou, Y., Zhao, B., Han, J., Zhou, W., Yao, H., and Xie, C.
\newblock How many unicorns are in this image? a safety evaluation benchmark for vision llms.
\newblock \emph{arXiv preprint arXiv:2311.16101}, 2023.

\bibitem[Vedantam et~al.(2015)Vedantam, Lawrence~Zitnick, and Parikh]{vedantam2015cider}
Vedantam, R., Lawrence~Zitnick, C., and Parikh, D.
\newblock Cider: Consensus-based image description evaluation.
\newblock In \emph{Proceedings of the IEEE conference on computer vision and pattern recognition}, pp.\  4566--4575, 2015.

\bibitem[Wang et~al.(2024)Wang, Zhou, Liu, Lu, Xu, He, Yoon, Lu, Bertasius, Bansal, et~al.]{wang2024mementos}
Wang, X., Zhou, Y., Liu, X., Lu, H., Xu, Y., He, F., Yoon, J., Lu, T., Bertasius, G., Bansal, M., et~al.
\newblock Mementos: A comprehensive benchmark for multimodal large language model reasoning over image sequences.
\newblock \emph{arXiv preprint arXiv:2401.10529}, 2024.

\bibitem[Wolf et~al.(2020)Wolf, Debut, Sanh, Chaumond, Delangue, Moi, Cistac, Rault, Louf, Funtowicz, et~al.]{wolf2020transformers}
Wolf, T., Debut, L., Sanh, V., Chaumond, J., Delangue, C., Moi, A., Cistac, P., Rault, T., Louf, R., Funtowicz, M., et~al.
\newblock Transformers: State-of-the-art natural language processing.
\newblock In \emph{Proceedings of the 2020 conference on empirical methods in natural language processing: system demonstrations}, pp.\  38--45, 2020.

\bibitem[Ye et~al.(2023)Ye, Xu, Ye, Yan, Liu, Qian, Zhang, Huang, and Zhou]{ye2023mplug}
Ye, Q., Xu, H., Ye, J., Yan, M., Liu, H., Qian, Q., Zhang, J., Huang, F., and Zhou, J.
\newblock mplug-owl2: Revolutionizing multi-modal large language model with modality collaboration.
\newblock \emph{arXiv preprint arXiv:2311.04257}, 2023.

\bibitem[Yin et~al.(2023)Yin, Fu, Zhao, Xu, Wang, Sui, Shen, Li, Sun, and Chen]{yin2023woodpecker}
Yin, S., Fu, C., Zhao, S., Xu, T., Wang, H., Sui, D., Shen, Y., Li, K., Sun, X., and Chen, E.
\newblock Woodpecker: Hallucination correction for multimodal large language models.
\newblock \emph{arXiv preprint arXiv:2310.16045}, 2023.

\bibitem[Zhai et~al.(2023)Zhai, Yang, Xu, Shen, Keutzer, and Li]{zhai2023halle}
Zhai, B., Yang, S., Xu, C., Shen, S., Keutzer, K., and Li, M.
\newblock Halle-switch: Controlling object hallucination in large vision language models.
\newblock \emph{arXiv e-prints}, pp.\  arXiv--2310, 2023.

\bibitem[Zhang et~al.(2024)Zhang, Zhao, Chen, Feng, Ding, and Sun]{zhang2024rankclip}
Zhang, Y., Zhao, Z., Chen, Z., Feng, Z., Ding, Z., and Sun, Y.
\newblock Rankclip: Ranking-consistent language-image pretraining.
\newblock \emph{arXiv preprint arXiv:2404.09387}, 2024.

\bibitem[Zhou et~al.(2023)Zhou, Cui, Yoon, Zhang, Deng, Finn, Bansal, and Yao]{zhou2023analyzing}
Zhou, Y., Cui, C., Yoon, J., Zhang, L., Deng, Z., Finn, C., Bansal, M., and Yao, H.
\newblock Analyzing and mitigating object hallucination in large vision-language models.
\newblock \emph{arXiv preprint arXiv:2310.00754}, 2023.

\bibitem[Zhou et~al.(2024{\natexlab{a}})Zhou, Cui, Rafailov, Finn, and Yao]{zhou2024aligning}
Zhou, Y., Cui, C., Rafailov, R., Finn, C., and Yao, H.
\newblock Aligning modalities in vision large language models via preference fine-tuning.
\newblock \emph{arXiv preprint arXiv:2402.11411}, 2024{\natexlab{a}}.

\bibitem[Zhou et~al.(2024{\natexlab{b}})Zhou, Fan, Cheng, Yang, Chen, Cui, Wang, Li, Zhang, and Yao]{zhou2024calibrated}
Zhou, Y., Fan, Z., Cheng, D., Yang, S., Chen, Z., Cui, C., Wang, X., Li, Y., Zhang, L., and Yao, H.
\newblock Calibrated self-rewarding vision language models.
\newblock \emph{arXiv preprint arXiv:2405.14622}, 2024{\natexlab{b}}.

\bibitem[Zhu et~al.(2023)Zhu, Chen, Shen, Li, and Elhoseiny]{zhu2023minigpt}
Zhu, D., Chen, J., Shen, X., Li, X., and Elhoseiny, M.
\newblock Minigpt-4: Enhancing vision-language understanding with advanced large language models.
\newblock \emph{arXiv preprint arXiv:2304.10592}, 2023.

\end{thebibliography}
\bibliographystyle{icml2024}

\newpage
\appendix
\onecolumn
\section{Proof of Robust Certification of FOV Sampling in
Theorem~\ref{theorem:robustness}}\label{app:robustness}
This section proves theoretical analysis on the robustness of HALC in approximating the 
optimal visual context $v^*$ via sampling in the FOV space 
(Theorem~\ref{theorem:robustness}). 
%
%
With certain assumptions on $v^*$ and $v_d$, we focus on demonstrating the certified robustness on the decoding token probability distribution compared with that from the optimal visual context $v^*$, when sampling different FOVs based on $v_d$ which is initially determined by an detector $\mathcal{G}_d$.

The objective of HALC is to approximate the unknown optimal visual context for a decoding step, thereby mitigating hallucination and enhancing the truthfulness of the LVLM outputs.
We approach the optimal proxy by sampling a series of $n$ FOVs in the original image $v$, starting from $v_d$ according to some sampling function $\pi(\cdot|v_d)$.
We focus on bounding the minimum deviation of the decoding token probabilities from the optimum among the $n$ FOV samples, with the hope that we can always find some sample that is close to the optimal $v^*$ during this process. And as the sample size $n$ becomes larger, the minimum deviation becomes smaller, indicating that we can better cover the optimal visual context $v^*$ within the samples.\footnote{The subsequent selection of the best sample is another question, which is not concerned in this proof. We theoretically justify the existence of an ``optimal'' sample in the proof here, and HALC selects such a sample by contrasting FOV pairs based on the observation illustrated in \figref{fig:optimal_visual_context}.}

%
\begin{proof}


Let $v^*=(w^*, h^*, p^*)$ be the optimal visual context, represented by a 3-tuple of its width, height, and center point.
The corresponding optimal token decoding probability distribution is $p_\theta(\cdot|v^*)$, where $\theta$ denotes the parameters of the LVLM {\lvlm}, and we ignore the condition on the textual query $x$ and previously generated tokens $y_{<t}$ for simplicity.
We rely on a symmetric discrepancy measure $D(\cdot,\cdot)\in [0, 1]$ to compare the disparity between two probability distributions, such as the Jensen-Shannon divergence, or the total variation distance.
We assume that the model prediction is robust around $v^*$ against small perturbations. In particular, we assume that there exists a tolerable small $\epsilon$-neighborhood $\mathcal{B}(v^*, \epsilon)=\{\hat{v}: \|\hat{v} - v^*\|\leq \epsilon\}$ around $v^*$, such that
\begin{equation}
    g(v^*, \hat{v}) = D(p_\theta(\cdot|v^*), p_\theta(\cdot|\hat{v})) \leq \delta\ll 1,\quad \forall \hat{v} \in \mathcal{B}(v^*, \epsilon)
\end{equation}
Essentially, for any visual context window (or FOV) close enough to $v^*$, the output token probability disparity is tiny, which is likely to result no difference in greedy decoding.

From the FOV detector $\mathcal{G}_d$, the output visual context is denoted as $v_d=(w_d, h_d, p_d)$, which is in general not the optimal.
We assume $v_d = v^* + \eta$ in the 3-tuple vector space, where $\eta$ is the perturbation vector from the optimal.
The detection perturbation is often large enough with $\|\eta\| > \epsilon$, making $v_d$ outside of the $\epsilon$-neighborhood of $v^*$.

$v_d \rightarrow v^*$:
If we directly use the detector output $v_d$ as an approximation of the optimal visual context $v^*$, the output distribution deviation from the optimum, measured by $g(v^*, v_d)$, is often unpredictable, when $v_d$ does not fall in the hypothetical tolerable region $\mathcal{B}(v^*, \epsilon)$.
An example can be seen as the inaccurate detection $v_d$ in \figref{fig:optimal_visual_context} results in the wrong token prediction \textit{book}.
This prompts the need for our proposed FOV sampling approach with the hope to find samples close to the optimal $v^*$.


\bfsection{$\pi(\cdot | v_d) \rightarrow v^*$:} 
Thus we consider sampling conditioned on $v_d$ in the FOV space to enhance the 
robustness of optimal visual context approximation, hoping to find some sample that is close to the optimal. To do this, we obtain an upper bound on the minimum deviation from the output distribution among a collection of FOV samples.
Assume $\pi(\cdot | v_d) \in \Omega$ is an arbitrary sampling function conditional on the initial FOV detection $v_d$, where $\Omega$ 
denotes the sampling space over all potential visual contexts in the entire image $v$. $\pi$ can either be a deterministic sampling function, or a stochastic sampling process with a probabilistic distribution over $\Omega$.
Suppose we acquire $n$ samples $v_1, v_2, \ldots, v_n$ according to $\pi(\cdot|v_d)$, we denote the minimum deviation of the resulted token probability from that of the optimal visual context $v^*$ as
\begin{equation}
    h_{\pi}(v^*, n) = \min_{i=1,\ldots, n} g(v^*, v_i)=\min_{i=1,\ldots, n} D\left(p_{\theta}(\cdot | v^*), p_{\theta}(\cdot | v_i)\right)
\end{equation}
where $D$ is the aforementioned symmetric discrepancy measure between two probability distributions, which is within the range of $[0, 1]$.
Having a small value of $h_{\pi}(v^*, n)$ would indicate that we can find some visual context that is close to the optimal $v^*$ through $n$ samples.

We proceed to estimate the minimum deviation $h_{\pi}(v^*, n)$ from the optimal visual 
context $v^*$ with $n$ samples.
We introduce a partition based on the occurrence 
of two probabilistic events: the event $A$ where at least one of the samples falls into the 
$\epsilon$-neighborhood $\mathcal{B}(v^*, \epsilon)$ close to $v^*$, and its complement. 
Let us denote the probability of at least one sample falling within $\mathcal{B}(v^*, \epsilon)$ as $P(A)$, and 
the complementary event's probability as $P(\neg A) = 1 - P(A)$. 
Hence, we can express the minimum divergence $h_{\pi}(v^*, n)$ as a marginalization over 
these events:
\begin{align}\label{eqn:expected_H}
h_{\pi}(v^*, n) &= P(A) \cdot [h_{\pi}(v^*, n) | A] + P(\neg A) \cdot [h_{\pi}(v^*, n) | \neg A]
\end{align}
Recognizing that for the one sample in the vicinity of $v^*$ in the event of $A$, its decoding token probability deviation from the optimal is bounded by $\delta\ll 1$ based on our assumption. Hence we have
\begin{align}\label{eq:general_bound}
    h_{\pi}(v^*, n) \leq & P(A) \cdot \delta + P(\neg A) \cdot 1 \leq  \delta + P(\neg A)
\end{align}
Next, we consider two instances of the sampling function $\pi(\cdot | v_d)$ that yield an upper bound for $h_{\pi}(v^*, n)$.

\bfsection{Normal Distribution Sampling.}
Suppose sampling from $\pi$ follows a stochastic process following a normal distribution around $v_d$. We denote this sampling process as $\pi_g(\cdot|v_d)\sim \mathcal{N}(v_d, \sigma^2 I)$, where we assume a variance of $\sigma^2$ for each element of the visual context representation (width, height, center) independently. 
For $\tilde{v}\in \Omega$, the probability of sampling $\tilde{v}$ following the multivariate normal distribution is 
\[q(\tilde{v}; v_d, \sigma^2 I) = \frac{1}{\sqrt{(2\pi\sigma^2)^s}} \exp\left(-\frac{1}{2\sigma^2}(\tilde{v} - v_d)^\top (\tilde{v} - v_d)\right) \]
where $s=3$ is the dimension of the FOV representation vector.
The probability of event $\neg A$ happening, which is none of $n$ FOV samples falling within the $\epsilon$-neighborhood of $v^*$, is
\begin{align}
    P(\neg A) &= P(\|v_1 - v^*\| > \epsilon) \wedge P(\|v_2 - v^*\| > \epsilon) \wedge \cdots P(\|v_n - v^*\| > \epsilon) \\
    &= P(\|\tilde{v} - v^*\| > \epsilon) ^ n \\
    &= P(\|\tilde{v} - (v_d - \eta)\| > \epsilon) ^ n 
\end{align}
%
From the normal distribution assumption of $\tilde{v}$, we know that $\tilde{v} - (v_d - \eta)$ also follows a normal distribution $\mathcal{N}(\eta, \sigma^2 I)$.
Therefore,
\begin{align}
    P(\neg A) &= \left(1 - P(\|\tilde{v} - (v_d - \eta)\| \leq \epsilon )\right) ^ n \\
    &= \left(1 - \int_{\nu: \|\nu\|\leq \epsilon} \frac{1}{\sqrt{(2\pi\sigma^2)^s}} \exp\left(-\frac{1}{2\sigma^2}(\nu - \eta)^\top (\nu - \eta)\right) d^s \nu \right) ^ n \\
    &= \left(1 - C_g(\epsilon, \eta;\sigma)\right) ^ n
\end{align}
where we use $C_g(\epsilon, \eta;\sigma)\in (0, 1)$ to denote the constant value given $\epsilon$, $\eta$, and $\sigma$.
Following \eqnref{eq:general_bound}, we now have
\begin{align}
    h_{\pi_g}(v^*, n) \leq \delta + (1 - C_g(\epsilon, \eta;\sigma))^n
\end{align}
where the second term goes to $0$ as $n$ is increasing to larger values.

\bfsection{Exponential Expansion Sampling.}
Now suppose sampling from $\pi$ follows an exponential expanding process, where a sample can be expressed as $v_r=(w_r, h_r, p_r)=((1+\lambda)^r w_d, (1+\lambda)^r h_d, p_d)$ with an expanding factor $\lambda$ (assuming $\lambda >0$ without loss of generality) and some $r$.\footnote{Besides expansion, this could also be an exponential shrinking process when $r$ is negative. We abuse the use of ``expansion'' for both.} Essentially, the sample space comprises all fields of view (FOVs) that maintain the same aspect ratio (i.e. $w_d/h_d$) and the same center $p_d$ with $v_d$.  
Assume the sampling is uniform among all possible FOVs in the sample space, which we denote as $\pi_e(\cdot|v_d)\sim \mathcal{U}(r\in [r_{\min}, r_{\max}])$, where $r_{\min}$ and $r_{\max}$ correspond to the smallest FOV allowed (such as a few pixels) and the largest FOV possible (i.e. the entire original image v), respectively.

For this sampling distribution, we introduce two moderate assumptions regarding the initial detection $v_d$.
First, the center of the detection is relatively close to the optimum, such that $|p_d - p^*|<\epsilon$.
Second, The detection $v_d$ and the optimum $v^*$ share the same aspect ratio, meaning $w_d/h_d=w^*/h^*$.
This assumption is reasonable since the optimum is unknown, and we can assume it adheres to the aspect ratio used by a standard detector.

We begin by deriving the range of $r$ such that $v_r$ falls into the small neighborhood $\mathcal{B}(v^*, \epsilon)$ around $v^*$. We need
\begin{align}
    &\|v_r - v^*\| \leq \epsilon \\
    \implies\quad (w_r - w^*)^2 + &(h_r - h^*)^2 + (p_r - p^*)^2 \leq \epsilon^2 \\
    \implies\quad [(1+\lambda)^r w_d - w^*]^2 + &[(1+\lambda)^r h_d  - h^*]^2 + (p_d - p^*)^2 \leq \epsilon^2 \\
    &\vdots \nonumber\\
    \implies\quad (w_d^2 + h_d^2)\left((1+\lambda)^r - \frac{w_d w^* + h_d h^*}{(w_d^2 + h_d^2)} \right)^2 &\leq \epsilon^2 - (p_d - p^*)^2 - \frac{h_d^2 {h^*}^2}{(w_d^2 + h_d^2)}(\frac{w_d}{h_d} - \frac{w^*}{h^*})^2 \\
    &= \epsilon^2 - (p_d - p^*)^2 >0
\end{align}
Denoting constants $C_a=\frac{\epsilon^2 - (p_d - p^*)^2}{(w_d^2 + h_d^2)}$ and $C_b=\frac{w_d w^* + h_d h^*}{(w_d^2 + h_d^2)}$, we get the range of $r$ such that $v_r\in \mathcal{B}(v^*, \epsilon)$ as
\begin{align}
    \max\left(r_{\min}, \frac{\log(C_b - \sqrt{C_a})}{\log(1+\lambda)}\right) &\leq r\leq \min\left(r_{\max}, \frac{\log(C_b + \sqrt{C_a})}{\log(1+\lambda)}\right)\quad\quad \text{if}\quad C_b > \sqrt{C_a} \\
    \text{Or}\hspace{1.5in} r_{\min} &\leq r\leq \min\left(r_{\max}, \frac{\log(C_b + \sqrt{C_a})}{\log(1+\lambda)}\right)\quad\quad \text{if}\quad C_b \leq \sqrt{C_a}
\end{align}
%
We further denote this range as $r\in[C_{\min}(\epsilon, v^*, v_d;\lambda), C_{\max}(\epsilon, v^*, v_d;\lambda)]$, with $r_{\min}\leq C_{\min}(\epsilon, v^*, v_d;\lambda) < C_{\max}(\epsilon, v^*, v_d;\lambda)\leq r_{\max}$.
Based on the independent uniform sampling assumption, the probability of the event $\neg A$ that none of the $n$ samples fall into the $\epsilon$-neighborhood around the optimum $\mathcal{B}(v^*, \epsilon)$ is
\begin{equation}
    P(\neg A) = \left(1-\frac{C_{\max}(\epsilon, v^*, v_d;\lambda) - C_{\min}(\epsilon, v^*, v_d;\lambda)}{r_{\max} - r_{\min}}\right)^n = \left(1-C_e(\epsilon, v^*, v_d;\lambda)\right)^n
\end{equation}
where we use $C_e(\epsilon, v^*, v_d;\lambda)\in(0, 1]$ to denote the constant value depending on $\epsilon, v^*, v_d, \lambda$.
Following \eqnref{eq:general_bound}, we then have
\begin{equation}
    h_{\pi_e}(v^*, n) \leq \delta + (1 - C_e(\epsilon, v^*, v_d;\lambda)))^n
\end{equation}
where the second term goes to 0 as $n$ is increasing to larger values.

\bfsection{Discussion.}
In the above, we demonstrated that beginning with the initial detected visual context $v_d$, under certain mild conditions, acquiring $n$ samples according to a distribution $\pi(\cdot|v_d)$ is an efficient method for identifying a sample that leads to a small bounded deviation in the token decoding probabilities from those derived from the optimal visual context $v^*$.
The more samples acquired, the tighter the bound is.  
This provides a simple and robust way of approximating the optimum.

Different sampling distributions have distinct characteristics.
For normal distribution sampling $\pi_g(\cdot|v_d)\sim\mathcal{N}(v_d, \sigma^2 I)$, the variance parameter $\sigma^2$ determines the spread of the samples and thus the likelihood of approximating the optimal $v^*$ within $\mathcal{B}(v^*, \epsilon)$.
For exponential expansion sampling $\pi_e(\cdot|v_d)\sim\mathcal{U}(r\in[r_{\min}, r_{\max}])$ with samples $v_r=((1+\lambda)^r w_d, (1+\lambda)^r h_d, p_d)$, the parameter $\lambda$ controls the rate of growth for the sampled visual contexts. In practice, we apply discrete integer values of $r$ to acquire different samples efficiently, thus $\lambda$ affects the sample coverage of the visual information around $v^*$.

The choice of the sampling distribution $\pi$ is contingent upon factors such as the quality of the detector $\mathcal{G}_d$, the LVLM backbone {\lvlm}, the textual query $x$, and the visual input $v$.
Specifically, the continuous normal distribution is advantageous for concentrated sampling around $v_d$, which is particularly effective when the detection perturbation $\eta$ is small (meaning $v_d$ is near $v^*$).
In contrast, exponential expansion sampling covers an extended range of visual contexts quickly, which is preferable when limited context information is obtained.
In scenarios where significant underestimation or overestimation in $G_d$ detection is present, the exponential expanding strategy can discover the optimal visual context more effectively. 
\end{proof}

\section{Time Cost Analysis}\label{app:time_cost}

Since time complexity is a critical aspect for VLM decoding algorithms, in this section we analyze the additional runtime overhead and time cost of HALC. According to \citet{biber2000longman}, nouns, adjectives, adverbs, numbers, verbs, and pronouns, which are tokens that will actually pass through HALC decoding, comprise approximately 35\% of the total words in modern English (we observe similar sparse patterns in our experiments). POS tagging is observably fast in practice (we used the spaCy package, which is highly optimized on CPU with the smallest tagger model, which is only 12 MB in size\footnote{\url{https://spacy.io/models/en\#en_core_web_sm}}). Thus we will mainly discuss the time cost w.r.t. other modules in HALC. 

For each individual token, after its original decoding, HALC will utilize the detection module to initialize the FOV sampling, for which we use $T_d$ to represent the detector time cost. Next, each one of the $n$ FOVs (in our experiments, $n=4$, as shown in~\tabref{tab:hyperparameter_halc}) is fed back into the LVLM for decoding, resulting in $n*T_{LVLM}$ time cost, where $T_{LVLM}$ represents the LVLM decoding time for a single step (although this may increase slightly as the sequence grows longer). Other computations on top of the multiple decodings such as contrasting the distributions can be ignored in comparison. Therefore, in summary, without any parallelization, for a sequence of $L$ tokens, HALC will cost approximately:
\begin{equation}
L * T_{LVLM} + L*0.35*(T_d + n*T_{LVLM})=L*((1 +0.35n)*T_{LVLM} + 0.35T_d)
\end{equation}
In practice, when $n=4$ and $T_d$ is relatively much smaller than $T_{LVLM}$ (the detection model Grounding DINO we used was based on the Swin-Tranformer\footnote{\url{https://huggingface.co/docs/transformers/model\_doc/swin}} with 341M parameters), we expect HALC to cost around 2.4x of the normal greedy decoding time expense. 

However, the decoding passes for the extra $n$ FOVs can essentially run \textbf{in parallel} as they do not depend on each other. With parallelization, the time cost with $n$ FOV decoding is equal to the time cost for 1 FOV decoding, so the expected time cost will be only approximately 1.35x of the greedy decoding. When the detection model time can not be ignored and in the worst case it is the same as the decoding step time (which is unlikely as the LVLMs we experimented with are 7B), the expected time cost would be 1.7x of the normal greedy decoding.

\section{Experimentation Details}\label{app:exp}
\subsection{Experimental Setups}\label{app:hyper}
The overall experiment settings is reported in \tabref{tab:hyperparameter_overall}. 
While the regular greedy decoding follows this setting, the beam search variant in our 
experiment essentially applies a token-wise beam search based on accumulated probability 
scores of the previous tokens $y_{<t}$. 
We use the default code for implementation of these two baselines in HuggingFace Transformers Repository~\cite{wolf2020transformers}.\footnote{\url{https://huggingface.co/docs/transformers}}

\begin{table}[H]
\centering
\caption{Overall Experiment Settings}
\begin{tabular}{l|c}
\hline
\textbf{Parameters} & \textbf{Value} \\ \hline
Maximum New Tokens (CHAIR) & $64$  \\ \hline
Maximum New Tokens (POPE)  & $64$  \\ \hline
Maximum New Tokens (MME)  & $128$ \\ \hline
Top-k & False \\ \hline
Top-p & $1$ \\ \hline
Temperature $\tau$ & $1$ \\ \hline
\end{tabular}
\label{tab:hyperparameter_overall}
\end{table}

The complete hyper-parameters for HALC in our experiments in \secref{sec:experiments} is 
reported in 
\tabref{tab:hyperparameter_halc}. Specifically, there are four major hyper-parameters that 
can actively adjust the effectiveness of HALC to adapt to different task settings:
\begin{enumerate}
    \item \textit{FOV Sampling Distribution}: Typically, a normal distribution, which concentrated around ${v}_d$, provides a tighter bound under minimal perturbations, while an exponential expansion sampling distribution, with a more averaged coverage of the sampling space, is preferable when less contexts of the task is available. Thus to preserve generality in our experiment, we have employed the exponential expansion sampling with exponential growth factor $\lambda=0.6$.
    \item \textit{Number of Sampled FOVs $n$}: $n$ determines the number of sampled FOVs in the sample space. According to Theorem~\ref{theorem:robustness}, while increasing $n$ and adjusting the distribution parameters can efficiently reduce minimum token probability deviations and enhance the robustness against perturbed initial detection, it's notable that the runtime costs also raise with $n$. Consequently, we set $n=4$ across all our experiments.
    \item \textit{JSD Buffer Size $m$}: For each beam in the overall beam search process (beam size $k$), our bi-adaptive visual grounding module samples $n$ visual contexts, which through interpolated JSD calculation would produce $\frac{n \cdot (n-1)}{2}$ JSD values in total. Then we select the top $m$ FOV pairs with relatively large discrepancy to produce contrastive candidate distributions.
    \item \textit{Beam Size $k$}: The beam size $k$ is set to adjust the diversity and range for HALC to search for the best candidate captions. Essentially, the global visual matching score module selects the top $k$ diverse captions from $2m \cdot k$ text sequence candidates passed from the local adaptive visual grounding module. While a larger $k$ involves a larger search space and hopefully a better generation, the runtime cost also raises linearly w.r.t. $k$. HALC adopts Bootstrapping Language-Image Pre-training (BLIP)~\cite{li2022blip} for both text and image encoding when computing their cosine similarity scores. Notably given the global search capability of our visual matching score module, HALC seeks to preserve a more diverse set of captions within the beam buffer.
    \item \textit{Other Hyperparameters}: Our implementation inherits an additional 
    hyperparameter, adaptive plausibility threshold, originally 
    from DoLA~\citep{chuang2023dola}. 
\end{enumerate}
%
%
\begin{table}[H]
\caption{HALC Hyperparameter Settings}
\centering
\begin{tabular}{l|c}
\hline
\textbf{Parameters} & \textbf{Value} \\ \hline
Amplification Factor $\alpha$ & $0.05$  \\ \hline
JSD Buffer Size $m$ & $6$  \\ \hline
Beam Size & $1$ \\ \hline
FOV Sampling & Exponential Expansion  \\ \hline
Number of Sampled FOVs $n$  & $4$  \\ \hline
Exponential Growth Factor $\lambda$ & 0.6\\ \hline
Adaptive Plausibility Threshold & $0.1$ \\
\hline
\end{tabular}
\label{tab:hyperparameter_halc}
\end{table}

Regarding the comparison of HALC with SOTAs that are specifically designed for OH mitigation, we 
adopt the code, hyper-parameters, and pre-trained models of each method outlined in their 
public repositories and papers respectively. 
Specifically, the hyper-paratermers for DoLa~\cite{chuang2023dola}\footnote{\url{https://github.com/voidism/DoLa}} 
is reported in \tabref{tab:hyperparameter_dola}; 
OPERA~\cite{huang2023opera}\footnote{\url{https://github.com/shikiw/OPERA}} 
is reported in \tabref{tab:hyperparameter_opera}; 
and the hyperparatermers for 
VCD~\cite{leng2023mitigating}\footnote{\url{https://github.com/DAMO-NLP-SG/VCD}} is 
reported in \tabref{tab:hyperparameter_vcd}. 
For each of these baselines, we strictly follow their implementations and default hyper-parameters 
as reported in the paper to reproduce their results. 
\begin{table}[H]
\caption{DoLa Hyperparameter Settings}
\centering
\begin{tabular}{l|c}
\hline
\textbf{Parameters} & \textbf{Value} \\ \hline
Repetition Penalty $\theta$ & $1.2$  \\ \hline
Adaptive Plausibility Threshold $\beta$ & $0.1$\\ \hline
Pre-mature Layers & $[0, 2 \cdots, 32]$\\
\hline
\end{tabular}
\label{tab:hyperparameter_dola}
\end{table}
\begin{table}[H]
\caption{OPERA Hyperparameter Settings}
\centering
\begin{tabular}{l|c}
\hline
\textbf{Parameters} & \textbf{Value} \\ \hline
Self-attention Weights Scale Factor $\theta$ & $50$  \\ \hline
Attending Retrospection Threshold & $15$\\\hline
Beam Size & $3$\\
\hline
Penalty Weights & $1$\\
\hline
\end{tabular}
\label{tab:hyperparameter_opera}
\end{table}
\begin{table}[H]
\caption{VCD Hyperparameter Settings}
\centering
\begin{tabular}{l|c}
\hline
\textbf{Parameters} & \textbf{Value} \\ \hline
Amplification Factor $\alpha$ & $1$  \\ \hline
Adaptive Plausibility Threshold & $0.1$\\\hline
Diffusion Noise Step & $500$\\
\hline
\end{tabular}
\label{tab:hyperparameter_vcd}
\end{table}

Regarding post-hoc correction method woodpecker~\cite{yin2023woodpecker}\footnote{\url{https://github.com/BradyFU/Woodpecker}}  and LURE~\cite{zhou2023analyzing}\footnote{\url{https://github.com/YiyangZhou/LURE}} , we also strictly follow their implementations and hyper-parameters 
as reported in the paper to reproduce their results. For woodpecker, we adopt their original code and use OpenAI API to access GPT-3.5 Turbo. In average, per 500 images would result in approximately \$4.5 cost. For LURE, we also directly adopt their pre-trained projection layer model (based on Minigpt4) to reproduce the results reported in this paper. All the hyper-parameters are default.

Notably, to construct a standardized evaluation platform, we reorganize these repositories and form a unified object hallucination evaluation benchmark released at~\url{https://github.com/BillChan226/HALC}. This benchmark repository provides at ease a unified access to most of the announced LVLMs for various VQA tasks, evaluated by CHAIR~\cite{rohrbach2018object} , POPE~\cite{li2023evaluating}, offline POPE (OPOPE), linguistic quality metrics and MME scores~\cite{fu2023mme} in a standardized pipeline.

\subsection{Empirical Studies on Optimal Visual Contexts}\label{app:contexts}
We verify our insight that optimal visual context is important in correcting object hallucination through an empirical pilot study. 
\figref{fig:pattern_analysis} shows the oracle performance of OH levels when we 
rely on optimal visual contexts for tokens through brute-force search, with greedy decoding 
on the MME benchmark~\cite{fu2023mme} on three categories of 
OH sources. Specifically, each MME sub-task contains 30 images, and we have followed~\cite{leng2023mitigating} and selected four sub-tasks (including \textit{existence}, \textit{count}, \textit{color}, \textit{position}) to evaluate the hallucination in our analysis, in total 110 distinct images. Based on these images, we manually constructed multiple challenging questions (2-4 per image) that are likely to induce the LVLM to hallucinate (e.g. queries based on co-occurrence statistics illustrated in~\cite{li2023evaluating} on some plausible but unfaithful objects that are likely to co-occur, some minor objects in the distance). Then we take each question as a count unit and calculate the number of hallucinations on word level (instead of token level) which could be attributed for each of the three sources. Then for each question with a hallucination occurring, we search across the original image input using a brutal-force breadth-first algorithms until the hallucinating token is corrected to be consistent with the ground truth. This process effectively succeeds to retrieve the optimal visual context for 54.0\% of the questions. For those questions that fail this brutal-force search, we further manually select the visual context candidates based on human priors. In total, 84.5\% of the questions that contain these three sources of hallucinations can be eliminated with an explicit optimal visual context $v^*$.

\section{MME Experiment Details}\label{app:mme}
The experiment details mostly follow Appendix~\ref{app:contexts}, where we adopt each sub-task of 30 images from the MME benchmark dataset\footnote{\url{https://github.com/BradyFU/Awesome-Multimodal-Large-Language-Models/tree/Evaluation}}, and reconstruct the question prompt following offline POPE. Specifically, instead of simply asking a question with a binary yes/no answer, we first ask the decoder to generate a detailed caption of the provided image and then check whether the target
positive/negative word existes in the caption. The detailed results are reported in Table~\ref{tab:MME_result}. The corresponding figure result is shown in~\figref{fig:mme_result}.

\begin{table}[h!]
\centering
\caption{Comparison of Decoder Performances on 4 MME sub-tasks}
\label{table:decoder_performance}
\begin{tabular}{@{}lcccccc@{}}
\toprule
Decoder & Existence & Position & Color & Count & Max Tokens & Num of Samples \\
\midrule
HALC  & 155 & 73.33 & 141.67 & 93.33 & 128 & 110 \\
Greedy & 145 & 63.33 & 118.33 & 85 & 128 & 110 \\
DoLa   & 145 & 60    & 118.33 & 85 & 128 & 110 \\
Opera  & 135 & 56.67 & 115    & 80 & 128 & 110 \\
VCD    & 135 & 70    & 133.33 & 70 & 128 & 110 \\
LURE   & 140 & 60    & 108.33 & 68.33 & 128 & 110 \\
\bottomrule
\label{tab:MME_result}
\end{tabular}
\end{table}

\newpage
\section{POPE Results}\label{app:pope}
Although we argue that POPE is not suitable for post-correction decoding methods and as 
a result we propose OPOPE. 
We also conduct the POPE evaluation and demonstrate the result in \tabref{tab:pope_results}. 
To adapt HALC for the original POPE benchmark, we use the \textbf{entire query} together with the initial 
answer (yes/no) as the text prompt for the detection model to provide a grounding for the 
focal area of the query.

\begin{table}[H]
\addtolength{\tabcolsep}{8pt}  
\renewcommand{\arraystretch}{1}  
\centering
\caption{Detailed POPE results with random, popular and adversarial samplings.}
\begin{tabular}{l l l l l l l}
\toprule
\textbf{Setting} & \textbf{Model} & \textbf{Decoding} & \textbf{Accuracy} & \textbf{Precision} & \textbf{Recall} & \textbf{$F_\text{1}$ Score} \\ 
\midrule
\multirow{5}{*}{Random} 
& \multirow{5}{*}{MiniGPT-4} & Greedy & 61.00 & 56.32 & 98.00 & 71.53 \\
&                          &Beam Search &58.00 &54.47 &97.33 &69.86 \\
&                          & OPERA   &57.66 &54.21 &98.67 &69.97 \\
&                          & VCD     &60.33 &57.87 &76.00 &65.71 \\
&                          & HALC    &61.33 &56.54 &98.00 &\textbf{71.70} \\ \cline{2-7}
\multirow{5}{*}{Popular} 
& \multirow{5}{*}{MiniGPT-4} & Greedy &55.33 &52.87 &98.00 &68.69 \\
&                          &Beam Search &50.33 &50.17 &97.33 &66.21 \\
&                          & OPERA   &51.00 &50.51 &98.67 &66.82 \\
&                          & VCD     &57.33 &55.05 &80.00 &65.21 \\
&                          & HALC    &55.67 &53.07 &98.00 &\textbf{68.85} \\ \cline{2-7}
\multirow{5}{*}{Adversarial} 
& \multirow{5}{*}{MiniGPT-4} & Greedy &54.00 &52.15 &96.7 &67.76 \\
&                          &Beam Search &52.00 &51.05 &97.33 &66.97 \\
&                          & OPERA   &52.67 &51.39 &98.67 &67.58 \\
&                          & VCD     &53.67 &52.53 &76.00 &62.13 \\
&                          & HALC    &56.00 &53.26 &98.00 &\textbf{69.02} \\
\bottomrule
\end{tabular}%
\label{tab:pope_results}
\end{table}
According to~\tabref{tab:pope_results}, HALC has also outperformed the other four methods by a large margin in terms of accuracy, precision, recall and F1 Score on all three types of POPE VQA tasks (random, popular, adversarial).

\newpage
\section{Comprehensive OPOPE Results}\label{app:opope}
\begin{table}[H]
\addtolength{\tabcolsep}{8pt}  
\renewcommand{\arraystretch}{1}  
\fontsize{8pt}{8pt}\selectfont
\centering
\caption{Detailed OPOPE results with random, popular and adversarial samplings.}
\begin{tabular}{l l l l l l l}
\toprule
\textbf{Setting} & \textbf{Model} & \textbf{Decoding} & \textbf{Accuracy} & \textbf{Precision} & \textbf{Recall} & \textbf{$F_\text{0.2}$ Score} \\ 
\midrule
\multirow{12}{*}{Random} 
& \multirow{6}{*}{MiniGPT-4} & Greedy & 68.30 & 97.24 & 37.67 & 91.67 \\
&                          &Beam Search & 68.37 & 96.30 & 38.20 & 90.98 \\
&                          & DoLa     & 68.50 & 97.27 & 38.07 & 91.78 \\
&                          & OPERA   & 68.67 & 96.98 & 38.53 & 91.63 \\
&                          & VCD     & 67.10 & 96.22 & 35.60 & 90.30 \\
&                          & Woodpecker   & 69.07 & 96.99 & 39.366 & 91.83\\
&                          & LURE   & 69.50 & 96.65 & 40.4 & 86.76 \\
&                          & HALC    & 67.90 & 97.36 & 40.4 & 91.74 \\ \cline{2-7}
& \multirow{6}{*}{LLaVA-1.5} & Greedy & 72.20 & 97.17 & 45.73 & 93.14 \\
&                          & Beam Search  & 71.33 & 97.48 & 43.80 & 93.09 \\
&                          & DoLa     & 72.30 & 96.78 & 46.13 & 92.86 \\
&                          & OPERA   & 71.20 & 96.76 & 43.87 & 92.47 \\
&                          & VCD     & 72.07 & 96.89 & 45.60 & 92.87 \\
&                          & Woodpecker   & 70.83 & 95.89 & 43.53 & 91.65\\
&                          & LURE   & 71.67 & 97.24 & 44.6 & 93.02\\
&                          & HALC    & 71.87 & 97.86 & 44.73 & 93.58 \\ \cline{2-7}
& \multirow{6}{*}{mPLUG-Owl2} & Greedy & 71.27 & 96.91 & 43.93 & 92.62 \\
&                          & Beam Search     & 70.50 & 97.26 & 42.20 & 92.61 \\
&                          & DoLa     & 71.47 & 96.92 & 44.33 & 92.69 \\
&                          & OPERA   & 70.17 & 96.92 & 41.67 & 92.22 \\
&                          & VCD     & 70.93 & 97.31 & 43.07 & 92.81 \\
&                          & Woodpecker   & 70.27 & 97.99 & 41.38 & 93.09\\
&                          & LURE   & 70.83 & 96.71 & 43.13 & 92.30 \\
&                          & HALC    & 71.50 & 97.38 & 44.20 & 93.07 \\ \cline{1-7}
\multirow{12}{*}{Popular} 
& \multirow{6}{*}{MiniGPT-4} & Greedy & 66.43 & 88.70 & 37.67 & 84.30 \\
&                          & Beam Search     & 67.00 & 90.09 & 38.20 & 85.62 \\
&                          & DoLa     & 66.8 & 89.50 & 38.07 & 85.08 \\
&                          & OPERA   & 66.80 & 88.65 & 38.53 & 84.43 \\
&                          & VCD     & 65.47 & 65.47 & 35.60 & 83.64 \\
&                          & Woodpecker   & 67.37 & 89.47 & 39.37 & 85.29 \\
&                          & LURE   & 67.8 & 89.38 & 40.4 & 85.40 \\
&                          & HALC    & 66.37 & 90.02 & 36.80 & 85.27 \\ \cline{2-7}
& \multirow{6}{*}{LLaVA-1.5} & Greedy & 70.27 & 89.79 & 45.73 & 86.58 \\
&                          & Beam Search  & 69.80 & 91.25 & 43.8 & 87.6 \\
&                          & DoLa     & 70.43 & 89.75 & 46.13 & 86.60 \\
&                          & OPERA   & 69.63 & 90.51 & 43.87 & 86.95 \\
&                          & VCD     & 70.57 & 91.08 & 45.60 & 87.71 \\
&                          & Woodpecker   & 69.37 & 90.07 & 43.53 & 86.51 \\
&                          & LURE   & 69.63 & 89.32 & 44.6 & 86.00 \\
&                          & HALC    & 70.03 & 90.74 & 44.67 & 87.28 \\ \cline{2-7}
& \multirow{4}{*}{mPLUG-Owl2} & Greedy & 69.30 & 89.13 & 43.93 & 85.74 \\
&                          & Beam Search     & 68.83 & 90.27 & 42.20 & 86.48 \\
&                          & DoLa     & 69.53 & 89.35 & 44.33 & 85.99 \\
&                          & OPERA   & 69.03 & 92.02 & 41.67 & 87.94 \\
&                          & VCD     & 69.43 & 91.10 & 43.07 & 87.35 \\
&                          & Woodpecker   & 68.58 & 90.73 & 41.38 & 86.75 \\
&                          & LURE   & 69.17 & 89.99 & 43.13 & 86.38 \\
&                          & HALC    & 69.63 & 89.95 & 44.20 & 86.50 \\ \cline{1-7}
\multirow{12}{*}{Adversarial} 
& \multirow{6}{*}{MiniGPT-4} & Greedy & 65.60 & 85.35 & 37.67 & 81.38 \\
&                          & Beam Search     & 66.3 & 87.21 & 38.20 & 83.11 \\
&                          & DoLa     & 65.87 & 85.74 & 38.07 & 81.80 \\
&                          & OPERA   & 66.3 & 86.66 & 38.53 & 82.68 \\
&                          & VCD     & 64.77 & 85.44 & 35.60 & 81.08 \\
&                          & Woodpecker   & 66.88 & 87.53 & 39.37 & 83.60 \\
&                          & LURE   & 67.13 & 86.82 & 40.4 & 83.14\\
&                          & HALC    & 66.00 & 88.47 & 36.80 & 83.94 \\ \cline{2-7}
& \multirow{6}{*}{LLaVA-1.5} & Greedy & 69.23 & 86.30 & 45.73 & 83.44 \\
&                          & Beam Search     & 68.47 & 86.45 & 43.8 & 83.33 \\
&                          & DoLa     & 69.33 & 86.07 & 46.13 & 83.30 \\
&                          & OPERA   & 68.37 & 86.01 & 43.87 & 82.95 \\
&                          & VCD     & 69.37 & 86.91 & 45.60 & 83.99 \\
&                          & Woodpecker   & 69.20 & 89.45 & 43.53 & 85.96 \\
&                          & LURE   & 68.7 & 86.1 & 44.6 & 83.13 \\
&                          & HALC    & 69.87 & 90.21 & 44.67 & 86.80 \\ \cline{2-7}
& \multirow{6}{*}{mPLUG-Owl2} & Greedy & 68.73 & 87.16 & 43.93 & 83.98 \\
&                          & Beam Search     & 68.27 & 88.17 & 42.20 & 84.63 \\
&                          & DoLa     & 68.87 & 87.02 & 44.33 & 83.91 \\
&                          & OPERA   & 68.57 & 90.22 & 41.67 & 86.35 \\
&                          & VCD     & 69.07 & 89.69 & 43.07 & 86.10 \\
&                          & Woodpecker   & 67.85 & 87.94 & 41.38 & 84.29\\
&                          & LURE   & 67.73 & 84.91 & 43.13 & 81.86\\
&                          & HALC    & 69.23 & 88.50 & 44.20 & 85.21 \\
\bottomrule
\end{tabular}%
\label{tab:opope_results}
\end{table}
%


\newpage
\section{Experiment Results on LLaVA-Bench}\label{app:llava_bench}
\begin{figure}[h]
    \centering
    \includegraphics[width=0.85\textwidth]{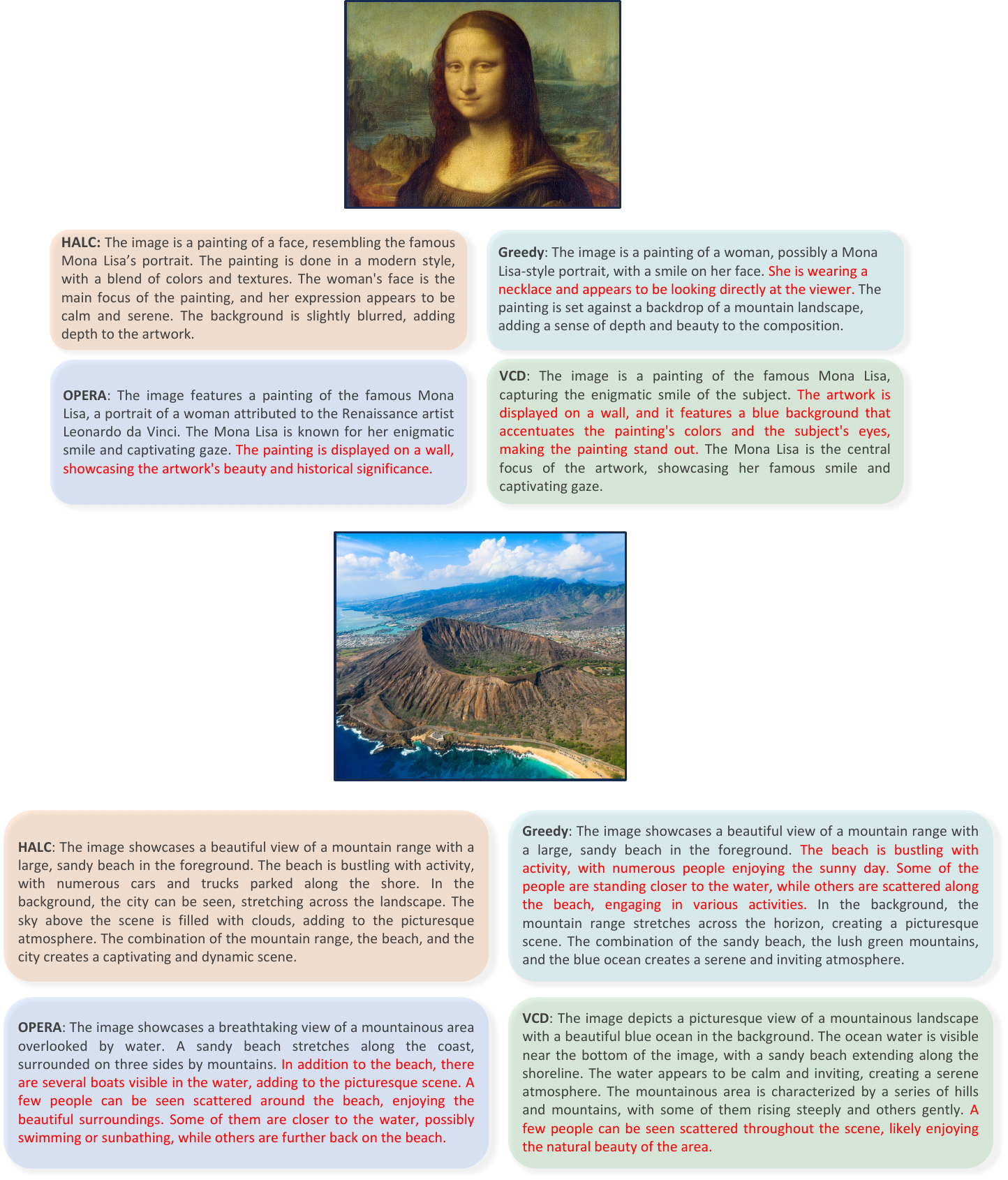}
    \caption{
        LLaVA-Bench results comparing HALC and other methods with mPLUG-Owl2~\cite{ye2023mplug}
        backbone.
    }
    \label{fig:llava_bench_mplug}
    \vspace{-0.15in}
\end{figure}

\begin{figure}[t]
    \centering
    \includegraphics[width=0.7\textwidth]{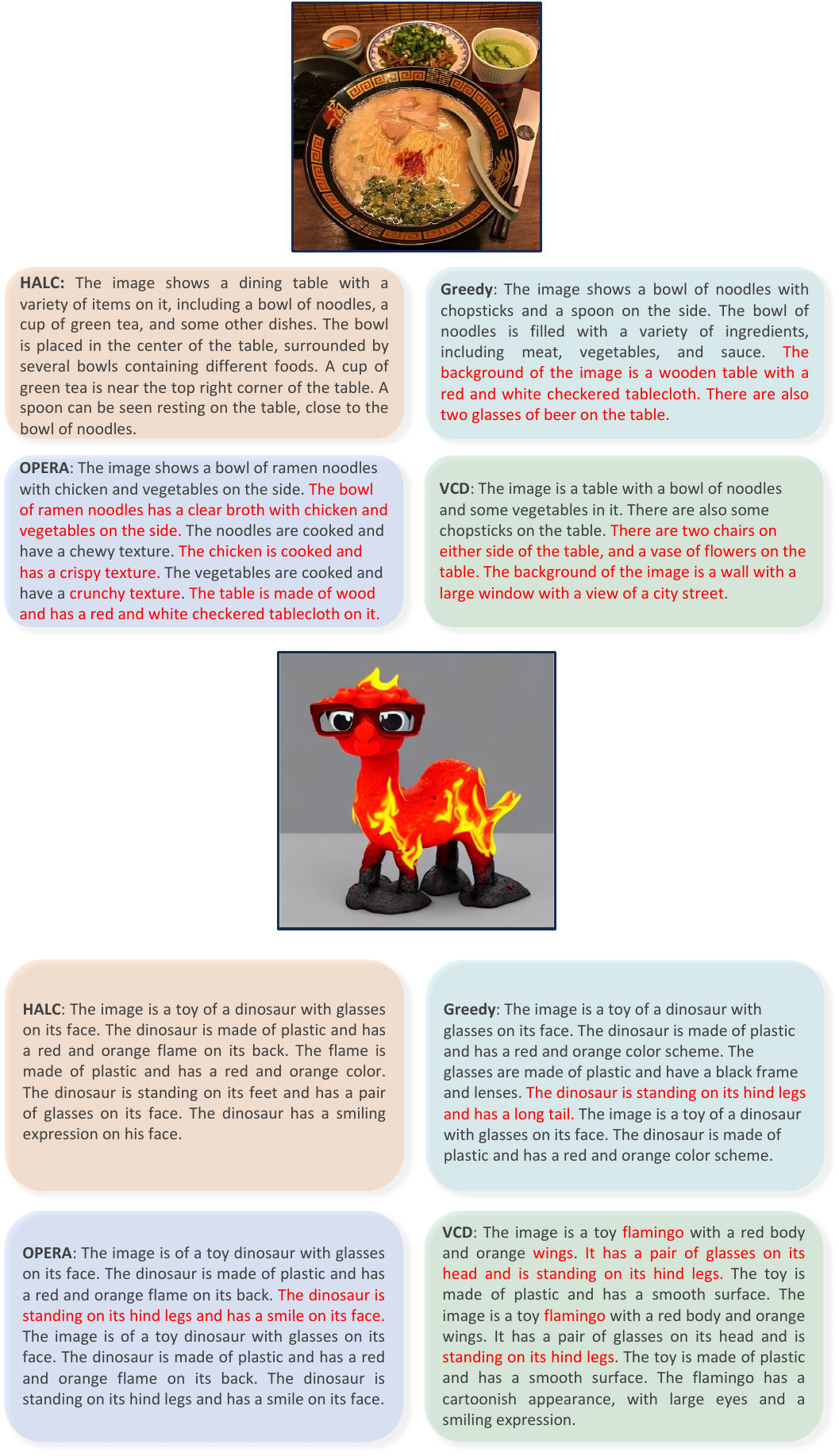}
    \caption{
        LLaVA-Bench results comparing HALC and other methods with 
        MiniGPT-4~\cite{zhu2023minigpt, chen2023minigpt} backbone.
    }
    \label{fig:llava_bench_minigpt4}
    \vspace{-0.15in}
\end{figure}

\begin{figure}[t]
    \centering
    \includegraphics[width=0.75\textwidth]{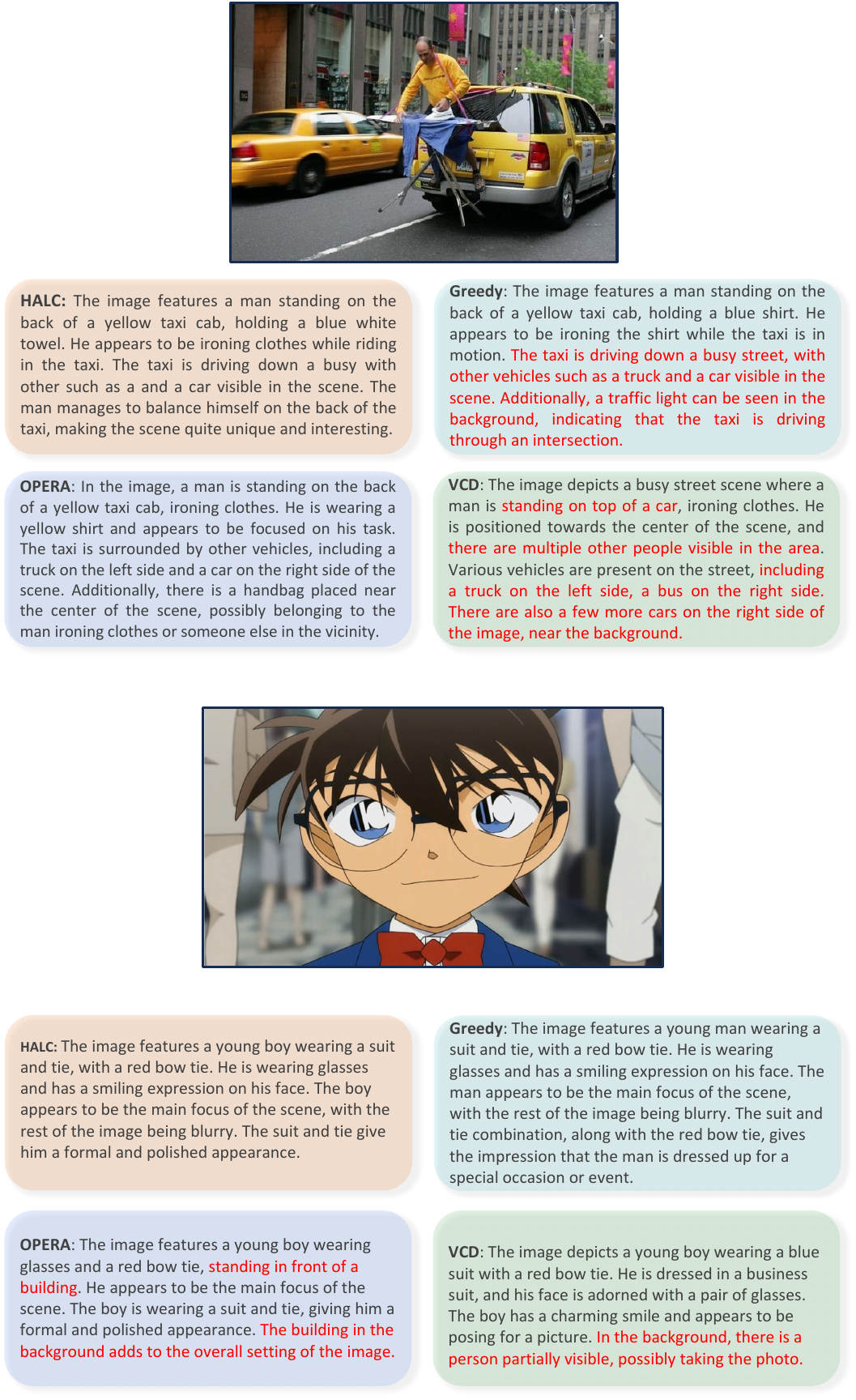}
    \caption{
        LLaVA-Bench results comparing HALC and other methods with LLaVA~\cite{liu2023visual}
        backbone.
    }
    \label{fig:llava_bench_llava}
    \vspace{-0.15in}
\end{figure}

As discussed in \secref{subsec:llava_bench}, we leverage LLaVA-Bench~\cite{liu2023improved} 
as a case study to qualitatively compare the decoding outputs of HALC with other methods.
Results generated by HALC and other OH reduction baselines incorporating 
mPLUG-Owl2~\cite{ye2023mplug}, MiniGPT-4~\cite{zhu2023minigpt, chen2023minigpt}, and
LLaVA~\cite{liu2023visual} LVLM backbones are shown in 
\figref{fig:llava_bench_mplug},~\ref{fig:llava_bench_minigpt4} and~\ref{fig:llava_bench_llava}
respectively.
In all the plots, red fonts indicate OH, including any of the object existence, attribute or
relationship hallucinations.

%




\end{document}